\documentclass{article}
\usepackage[utf8]{inputenc}
\usepackage[T1]{fontenc}
\usepackage{times}
\usepackage[margin=1in]{geometry}
\usepackage{natbib}
\setcitestyle{authoryear,round,citesep={;},aysep={,},yysep={;}}
\usepackage{hyperref,url}
\usepackage{amsmath,amssymb,amsthm}
\usepackage{graphicx}
\usepackage{xcolor,microtype,booktabs,placeins}
\usepackage{authblk}

\newtheorem{proposition}{Proposition}
\title{DuetMoE: Coupling Inter- and Intra-Subgroup\\Robustness for Fair Medical Image Analysis}
\author[1,2]{Yiqi Tian\textsuperscript{*}}
\author[5]{Jinwoong Park\textsuperscript{*}}
\author[3,4]{Sangjoon Park\textsuperscript{*}}
\author[1]{Pengfei Jin}
\author[2]{Bo Zeng}
\author[4,5]{Yujin Oh\textsuperscript{\dag}}
\author[1]{Quanzheng Li\textsuperscript{\dag}}
\affil[1]{Center for Advanced Medical Computing and Analysis, Massachusetts General Hospital and Harvard Medical School, Boston, MA 02114}
\affil[2]{Department of Industrial Engineering, University of Pittsburgh, Pittsburgh, PA 15261}
\affil[3]{Department of Radiation Oncology, College of Medicine, Yonsei University, Seoul, South Korea}
\affil[4]{Institute for Innovation in Digital Healthcare, Yonsei University, Seoul, South Korea}
\affil[5]{Department of Biomedical Systems Informatics, College of Medicine, Yonsei University, Seoul, South Korea}
\affil[ ]{\normalfont\small\textsuperscript{*}Equal contribution.}
\affil[ ]{\normalfont\small\textsuperscript{\dag}Co-corresponding authors: 
\texttt{yujinoh@yuhs.ac}, \texttt{li.quanzheng@mgh.harvard.edu}.}
\date{}
\hypersetup{hidelinks,pdfauthor={Yiqi Tian, Sangjoon Park, Bo Zeng, Pengfei Jin, Yujin Oh, Quanzheng Li},pdftitle={DuetMoE: Coupling Inter- and Intra-Subgroup Robustness for Fair Medical Image Analysis}}
\begin{document}
\maketitle

\begin{abstract}
As medical AI expands across diverse healthcare settings worldwide, equitable performance across patient populations is becoming essential to trustworthy clinical use. Fairness in medical image analysis is often evaluated through average performance across predefined subgroups, yet similar subgroup averages can conceal substantial variation among individual patients. Therefore, a reliable medical AI requires addressing two complementary objectives: \emph{inter-subgroup fairness}, which reduces performance disparities across groups, and \emph{intra-subgroup robustness}, which protects poorly served patients within each group. To jointly address these objectives, we propose \textbf{DuetMoE}, a subgroup-aware mixture-of-experts framework that couples group-level adaptation with patient-specific clinical guidance, enabling more reliable medical image analysis for individual patients. For settings without linked clinical records, we further introduce \textbf{DuetMoE+}, which retains subgroup-routed experts and incorporates a KL-constrained distributionally robust objective to address intra-subgroup robustness. We evaluate our methods on PI-CAI, radiotherapy, and Harvard-FairSeg. Across four evaluation settings, our methods lead in overall or equity-scaled performance, reduce the inter-subgroup mean Dice gap by up to $30.7\%$, and raise intra-subgroup 25th-percentile Dice by up to 11.5 points over the strongest reported baselines.
\end{abstract}

\section{Introduction}
\label{sec:intro}

Medical image analysis supports disease detection and the delineation of structures used in assessment and treatment planning, from prostate MRI to optic structures in glaucoma evaluation and targets in radiotherapy~\citep{saha2024picai,tianfairseg,choi2020clinical}. Missed lesions or inaccurate boundaries can change the assessment of disease or the extent of a planned treatment target. Reliable predictions therefore matter at the level of the patient whose care they inform. Yet patients within the same institution, tumor-stage group, or demographic group can differ in anatomy, disease extent, and clinical history, and a strong subgroup average may still conceal poor predictions for some of them. This motivates two complementary goals: inter-subgroup fairness, addressing performance disparities across groups, and intra-subgroup fairness, addressing patient-level reliability within each group. Progress on one does not establish progress on the other. Addressing both requires learning subgroup commonalities while accounting for individual differences~\citep{liu2023translational}.

Clinical records available at the relevant care stage may include age, laboratory measurements such as organ-specific antigen levels, tumor stage, and surgical history~\citep{saha2024picai}. Multimodal artificial intelligence (AI) integrates this context with imaging to support patient-specific segmentation and downstream decisions that better reflect clinical reasoning. Recent approaches use large language models (LLMs) to encode heterogeneous clinical information alongside visual representations~\citep{xiao2026medicalmllm}. Segmentation models align LLM-derived clinical representations with image features~\citep{oh2023llm,rajendran2025autodelineation,huemann2026contextual}, while MoME adds center-specific expert routing to address institutional bias~\citep{oh2024mixture}. Although these methods personalize predictions using clinical context, their architectures do not explicitly tailor computational pathways or clinical guidance to patient-level differences within the same subgroup.

Patients within the same subgroup can differ in both their clinical circumstances and the image features that reach a shared expert. The useful clinical guidance can therefore differ even when the router and expert are shared. To address both levels, we retain subgroup-specific routing while letting each patient's clinical context shape expert selection and the guidance each expert receives after routing. This motivates us to propose \textbf{DuetMoE}. After routing, each expert retrieves \textit{expert-specific clinical guidance} using the features it actually receives. By matching this guidance to each patient's circumstances and assigned features, the design targets intra-subgroup robustness. 

Beyond these information-rich settings, some clinical workflows provide only a single task modality without linked clinical records. However, reliable predictions are still needed both across subgroups and among patients within each group. When subgroup labels are available but linked clinical records are not, we introduce \textbf{DuetMoE+}, a unimodal variant that retains subgroup-routed experts and uses within-subgroup distributionally robust optimization (DRO) to emphasize high-loss patients.
\begin{figure}[t!]
    \centering
    \includegraphics[width=0.85\linewidth]{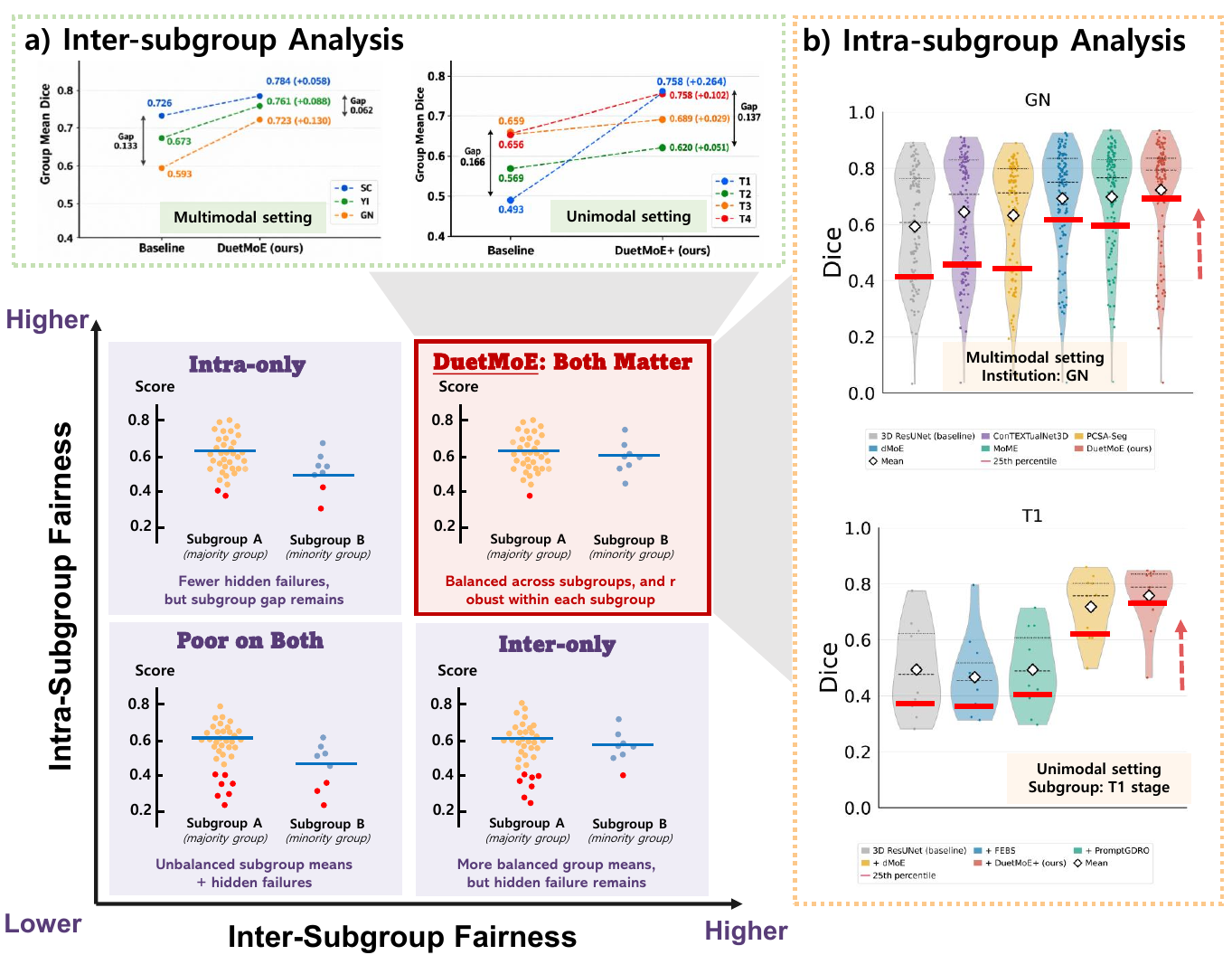}
    \caption{\textbf{Overview}. The central schematic depicts inter-subgroup balance and intra-subgroup patient outcomes as complementary goals. The surrounding panels show results from our proposed methods: (a) at the \textit{inter-subgroup level}, subgroup mean performance improves and becomes more balanced; and (b) at the \textit{intra-subgroup level}, the 25th percentile of patient-level Dice increases.}
    \label{fig:overview}
\end{figure}
Figure \ref{fig:overview} summarizes the inter- and intra-subgroup perspective behind \textbf{DuetMoE} and illustrates our methods’ performance on both axes. Our contributions are threefold:
\begin{itemize}
     \item \textbf{DuetMoE: inter-subgroup routing and intra-subgroup clinical guidance.} We use subgroup-specific routers to adapt across groups and embed patient-specific clinical memory before and after routing, tailoring expert selection and matching each selected expert's guidance to its assigned image features.
    \item \textbf{DuetMoE+: inter-subgroup routing and intra-subgroup robust training.} Without linked clinical records, the same subgroup label selects the router and defines a within-subgroup KL-robust objective that gives high-loss patients greater training influence.
    \item \textbf{Computational results on both fairness axes.} We conduct extensive comparisons on two public benchmarks, PI-CAI and Harvard-FairSeg, and an in-house radiotherapy dataset, evaluating both subgroup performance and patient-level outcomes. Across four settings, our methods lead in overall or equity-scaled performance. For inter-subgroup balance, our methods reduce the inter-institution mean Dice gap by up to $30.7\%$ relative to the best prior model in overall Dice. For intra-subgroup robustness, our methods improve lowest-$10\%$ mean cup Dice by up to $8.4$ percentage points over the subgroup-routing baseline.
\end{itemize}

\section{Related Work}

\paragraph{Fairness across and within subgroups.}
Fairness methods in medical segmentation act on data, representations, and training objectives. FairSeg combines demographic subgroup evaluation with Fair Error-Bound Scaling motivated by difficult cases within groups~\citep{tianfairseg}; FairDiff improves group representation through synthetic data~\citep{li2024fairdiff}, and FairDomain incorporates demographic information into cross-domain feature learning~\citep{tian2024fairdomain}. Hidden stratification can leave clinically important failures obscured by aggregate performance~\citep{oakdenrayner2020hidden}. Even when a method addresses difficult cases during training, subgroup means alone do not reveal how outcomes vary among patients in that subgroup. This motivates considering inter-subgroup disparities and intra-subgroup patient variation as complementary evaluation axes.

\paragraph{Clinical context in multimodal medical image analysis.}
Multimodal segmentation uses clinical information to contextualize image features. In routine clinical workflows, disease-stage information and patient-level variables, such as demographics, biomarkers, and organ-specific measurements, are often available before treatment planning or medical image interpretation. For radiotherapy planning, LLMSeg and ConTEXTualNet3D combine language-model-based clinical encoding with layer-wise image--text alignment~\citep{oh2023llm, huemann2026contextual}, while Medformer employs vision--language attention and language gating across multiple visual stages~\citep{rajendran2025autodelineation}. PCSA-Seg maintains language guidance throughout hierarchical decoding using clinical text available at image analysis~\citep{XUE2026115745}. Together, these methods show how clinical information can guide feature processing across scales and stages to better align AI predictions with clinical practice.

\paragraph{Subgroup-aware and context-conditioned experts.}
In dMoE, a subgroup attribute selects a router that directs image features to shared experts~\citep{oh2025distribution}, while LTSE uses image features and text describing a requested tissue target to select expert decoders for pathology referring segmentation~\citep{tang2025ltse}. Relatedly, the Mixture of Multicenter Experts (MoME) combines clinical fusion with center-specific expert routing~\citep{oh2024mixture}. In MoME, however, clinical alignment is applied after the expert mixture, rather than retrieving clinical guidance conditioned on the features assigned to each expert. DuetMoE instead integrates clinical context directly into subgroup-aware expert computation. 

\paragraph{Intra-subgroup robust risk.}
When auxiliary clinical context is unavailable but subgroup labels are known, loss-based objectives can give greater training weight to high-loss patients within each group. GroupDRO targets the worst group-average loss~\citep{sagawa2019distributionally}, while risk-averse adaptive sampling and positive loss tilting give more influence to difficult examples~\citep{curi2020adaptive,li2023tilted}. Hierarchical tilted risk and hierarchical ambiguity sets also account for variation both across and within groups~\citep{li2023tilted,jo2026hierarchical}. These general-purpose objectives provide useful tools for intra-subgroup risk control.

\section{Method}
\label{sec:method}
\suppressfloats[t]

To address both inter- and intra-subgroup fairness, we propose \textbf{DuetMoE}, which combines subgroup-selected routing over shared experts with patient- and region-specific clinical context to guide routing and expert computation. When linked clinical records are unavailable, its unimodal variant \textbf{DuetMoE+} retains subgroup routing and adds a within-subgroup robust loss that emphasizes higher-loss patients. 

\begin{figure}[!t]
    \centering
    \includegraphics[width=\linewidth]{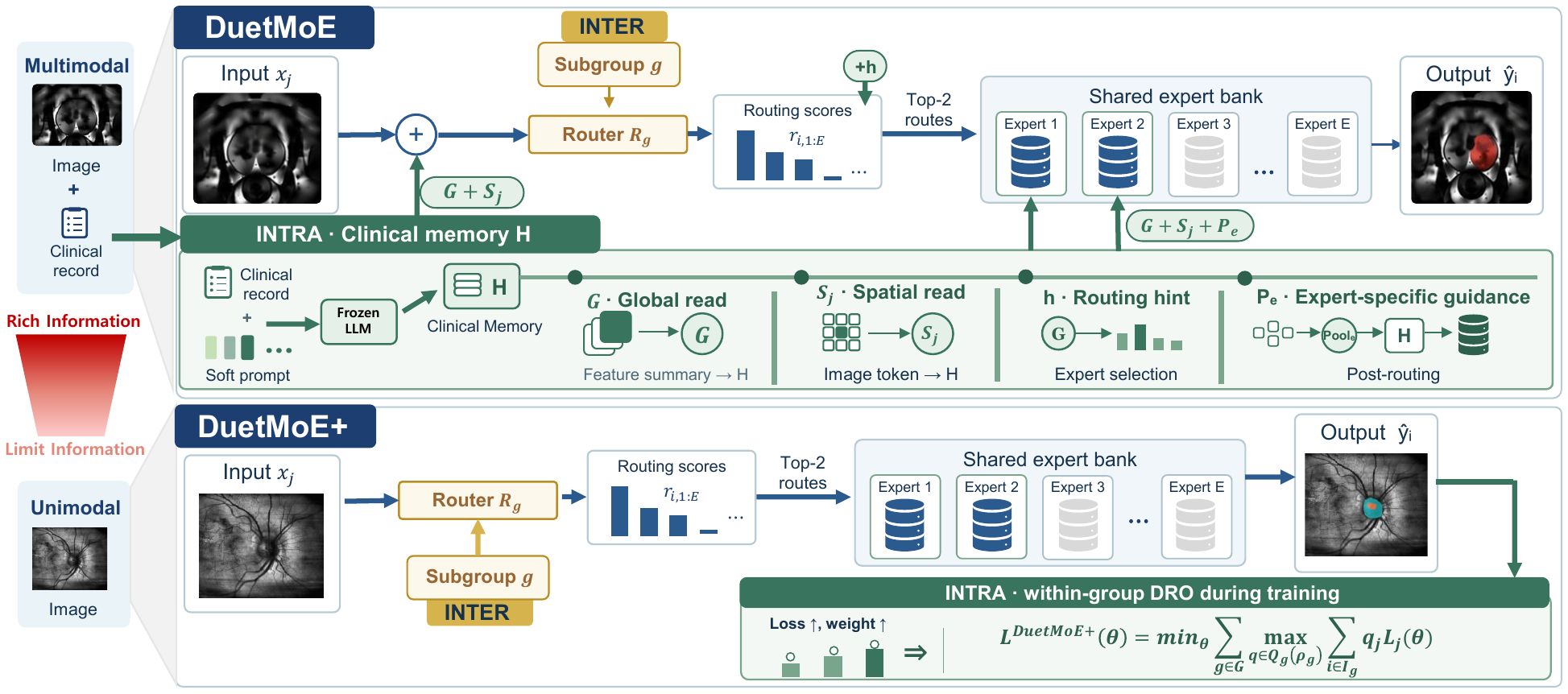}
    \caption{Summary of \textbf{DuetMoE} and \textbf{DuetMoE+}. Inter-subgroup adaptation: both variants use subgroup-specific routers to select shared experts. Intra-subgroup robustness: \textbf{DuetMoE} (top) embeds patient-specific clinical memory before and after routing, using global and spatial reads to guide expert selection and assigned image features to retrieve expert-specific guidance. Without linked clinical records, \textbf{DuetMoE+} (bottom) uses within-subgroup KL-DRO to give higher-loss patients greater influence on training their subgroup router and the shared experts.}
    \label{fig:method_overview}
\end{figure}

Figure~\ref{fig:method_overview} summarizes both settings. For each of $N$ patients, let $x_j$ denote the task input, $y_j$ the segmentation target, $g_j\in\mathcal G$ the subgroup label, and $c_j$ the optional clinical context. We write $L_j(\theta)$ for the patient segmentation loss under model parameters $\theta$. 

\subsection{{DuetMoE: adapting experts to groups and patients}}
\label{sec:duetmoe}

A subgroup label alone cannot tell the model which clinical information matters for a particular patient's image. The same clinical variable may matter differently when the model uses a summary of the current feature map or features at individual locations; a shared expert may also receive different feature patterns from patients in that subgroup. Rather than assigning each variable a fixed role, \textbf{DuetMoE} lets both image views query the patient's clinical memory before routing and each expert's assigned features query it afterward. This aims to preserve inter-subgroup differences via subgroup-specific routing, while achieving intra-subgroup robustness through patient-conditioned reads that guide expert selection and computation.

\paragraph{Global and spatial clinical context.}
For patient $j$, a frozen language model with learnable soft prompts produces clinical memory $H_j=\{h_{jm}\}_{m=1}^{M_j}$ with $M_j$ valid vectors~\citep{lester2021power}. At an encoder stage (index omitted), let $v_{jt}$, $t=1,\ldots,T$, be the tokens at its $T$ image locations. Global and spatial queries read $H_j$ using a summary of the current feature map and features at individual locations. Define $\hat v_{jt}=\operatorname{LN}(v_{jt})$ and $\bar v_j=T^{-1}\sum_t\hat v_{jt}$ as the normalized tokens and their mean, with learned layer normalization $\operatorname{LN}$ over token channels. Let $Q$ and $O$ be the query and output projections. The multi-head attention output $\mathcal A(q,H_j)$ uses projected query $q$ and keys and values from $H_j$, before projection by $O$. Writing $A_j^G$ for the global attention output before projection, the global context $G_j$ and spatial context $S_{jt}$ are
\begin{equation}
\begin{aligned}
A_j^G&={\mathcal A\!\left(Q(\bar v_j),H_j\right),\quad G_j=O(A_j^G),} \quad
{S_{jt}}=O\!\left(\mathcal A\!\left(Q(\hat v_{jt}),H_j\right)\right).
\end{aligned}
\label{eq:duetmoe_context}
\end{equation}

\paragraph{Routing with clinical context.}
The subgroup label $g_j$ selects the router $R_{g_j}$, which scores $E$ shared experts using image tokens augmented with global and spatial context. A separate linear projection $B$, shared across subgroups, maps the global attention output $A_j^G$ to a routing hint for each expert: $u_{jt}$ is the augmented router input, $r_{jt}\in\mathbb R^E$ the expert-score vector, and $\pi_{jt}\in\mathbb R^E$ the resulting sparse weights.
\begin{equation}
\begin{aligned}
u_{jt} =v_{jt}+\alpha G_j+\beta {S_{jt}}, \quad r_{jt}=R_{g_j}(u_{jt})+\lambda B(A_j^G)+\xi_{jt}, \quad
\pi_{jt}=\operatorname{SoftmaxTopK}(r_{jt}).
\end{aligned}
\label{eq:duetmoe_router}
\end{equation}

During training and evaluation, $\xi_{jt}\in\mathbb R^E$ is the score-noise vector in noisy top-$K$ gating~\citep{shazeer2017outrageously}. $\operatorname{SoftmaxTopK}$ softmax-normalizes the $K$ largest scores (expert set $\mathcal K_{jt}$) and sets the rest to zero. The component $\pi_{jt,e}$ is the weight of expert $e$.

\paragraph{Expert-specific clinical guidance after routing.}
After routing, we summarize the normalized image features assigned to each expert and use them to read the clinical memory again. {Here $z_{j,e}$ is the routing-weighted summary and $P_{j,e}$ the guidance read for patient $j$ and expert $e$.} For each patient--expert pair satisfying $\sum_t\pi_{jt,e}>0$, we define
\begin{equation}
\begin{aligned}
z_{j,e}=\frac{\sum_t\pi_{jt,e}\hat v_{jt}}{\sum_t\pi_{jt,e}}, \quad
P_{j,e}=O\!\left(\mathcal A\!\left(Q(z_{j,e})+b_e,H_j\right)\right),
\end{aligned}
\label{eq:duetmoe_guidance_main}
\end{equation}
where $b_e$ is a learnable query offset for expert $e$. The summary $z_{j,e}$ is computed separately for each patient, so the clinical guidance $P_{j,e}$ can differ between patients who share a router and expert. The selected experts $E_e$ process tokens augmented with $G_j$, $S_{jt}$, and $P_{j,e}$; $\widetilde v_{jt,e}$ denotes that expert's input and $v_{jt}^{\mathrm{out}}$ the updated token. Their outputs form a weighted residual update:
\begin{equation}
\begin{aligned}
\widetilde v_{jt,e}=v_{jt}+\gamma G_j+\delta {S_{jt}}+\varepsilon P_{j,e}, \quad
v_{jt}^{\mathrm{out}}=v_{jt}
+\sum_{e\in\mathcal K_{jt}}\pi_{jt,e}E_e\!\left(\widetilde v_{jt,e}\right).
\end{aligned}
\label{eq:duetmoe_output}
\end{equation}

\paragraph{Architecture and training.} 
Only the last four of five encoder stages use clinical context. The six coefficients $(\alpha,\beta,\lambda,\gamma,\delta,\varepsilon)$ are learned separately at each clinical-context stage and shared across patients. More implementation details are provided in Appendix~\ref{app:duetmoe_details}.

\subsection{DuetMoE+: inter-subgroup routing and intra-subgroup robust risk}
\label{sec:duetmoeplus}

Clinical records may be unavailable or unlinked even when images and subgroup labels are available, as in fundus datasets with demographic labels or radiotherapy archives with planning CT scans and tumor-stage labels. These settings lack the patient-specific clinical memory needed for \textbf{DuetMoE}'s context-guided routing and expert updates. To address both subgroup disparities and unreliable predictions within groups, we introduce \textbf{DuetMoE+}, a variant that retains subgroup-selected expert routing and targets intra-subgroup fairness through a robust intra-subgroup loss.

\paragraph{\textbf{Architecture-level inter-subgroup adaptation.}}
We construct \textbf{DuetMoE+} from \textbf{DuetMoE} by first removing the clinical pathways used before and after routing: the global and spatial reads, the clinical routing hint, and the expert-specific read. We keep the subgroup-selected routers and shared experts, so the available subgroup label can still shape how image features are processed. In such unimodal setting, the inter-subgroup architecture of \textbf{DuetMoE+} reduces to dMoE, as shown in Proposition \ref{prop:duetmoe_context_free}, with detailed proof in Appendix~\ref{app:duetmoeplus_details}. 

\begin{proposition}[Inter-subgroup architecture reduces to dMoE~\citep{oh2025distribution}]
\label{prop:duetmoe_context_free}
At an encoder stage, disabling all clinical-memory reads and the memory-derived routing hint in DuetMoE recovers the dMoE update for every patient and image token, with the subgroup routers, shared experts, and gating rule held fixed.
\end{proposition}

\paragraph{\textbf{Loss-level intra-subgroup robust risk.}}
A subgroup mean can hide high-loss patients. \textbf{DuetMoE+} uses within-subgroup KL-DRO to give these patients greater training influence on the subgroup routers and shared experts. Let $I_g=\{j:g_j=g\}$ and $\widehat P_g$ be the uniform empirical distribution on $I_g$. For $\rho_g\geq0$, define $\mathcal U_g^{\mathrm{KL}}(\widehat P_g) =\{Q_g:D_{\mathrm{KL}}(Q_g\Vert\widehat P_g)\leq\rho_g\}$, where $Q_g$ ranges over distributions on $I_g$. We train by
\begin{equation}
\min_\theta\ \mathcal L_{\mathrm{DuetMoE+}}(\theta)
=\min_\theta\sum_{g\in\mathcal G}\widehat R_g^{\mathrm{rob}}(\theta), \quad \widehat R_g^{\mathrm{rob}}(\theta)=\sup_{Q_g\in\mathcal U_g^{\mathrm{KL}}(\widehat P_g)}\mathbb E_{j\sim Q_g}\!\left[L_j(\theta)\right].
\label{eq:duetmoeplus_group_risk}
\end{equation}
The KL radius controls departures from uniform patient weights, allowing greater emphasis on high-loss observed patients. When $\rho_g=0$ for all groups, the objective reduces to the sum of subgroup mean losses. The next proposition relates this robust risk to within-subgroup loss variation. The following result shows how the robust risk responds to variation among patients within a subgroup.

\begin{proposition}[Sensitivity to intra-subgroup variation]
\label{prop:duetmoeplus_variance}
Fix a nonempty subgroup $g$ with finite patient losses. Let $\widehat R_g^{\mathrm{mean}}(\theta)=|I_g|^{-1}\sum_{j\in I_g}L_j(\theta)$ and $\sigma_g(\theta)$ be their mean and standard deviation. Replacing each robust risk in Eq.~\eqref{eq:duetmoeplus_group_risk} by its mean defines the mean-loss (ERM) version, recovered when all $\rho_g=0$. For fixed $\theta$, as $\rho_g\downarrow0$,
\begin{equation}
\widehat R_g^{\mathrm{rob}}(\theta)
=\underbrace{\widehat R_g^{\mathrm{mean}}(\theta)}_{\text{mean-loss version}}
+\underbrace{\sqrt{2\rho_g}\,\sigma_g(\theta)}_{\text{intra-subgroup variation penalty}}
+O(\rho_g).
\label{eq:duetmoeplus_variance}
\end{equation}
If $\sigma_g(\theta)=0$, robust and mean risks coincide for every $\rho_g\geq0$.
\end{proposition}

A low subgroup mean loss can hide large errors on individual patients. Proposition~\ref{prop:duetmoeplus_variance} shows that, for small $\rho_g$, the robust loss approximately equals the mean loss plus a penalty proportional to the standard deviation of patient losses, making intra-subgroup variation explicit in the training objective. The KL weights give high-loss patients greater training influence within their subgroup; Appendix~\ref{app:duetmoeplus_details} provides the proof and weight derivation.

\section{Experiments}
\label{sec:experiments}
Our evaluation follows the information available to the predictor. We evaluate \textbf{DuetMoE} on PI-CAI and institution-grouped radiotherapy (RT), where images are linked to clinical records, and its unimodal variant \textbf{DuetMoE+} on T-stage-grouped RT and Harvard-FairSeg, where images and subgroup labels are available without additional clinical context. Subgroup results assess inter-subgroup performance, while available patient-level results examine intra-subgroup reliability. The two RT protocols use different cohorts and subgroup definitions and are analyzed separately.

\subsection{Datasets, baselines, and evaluation metrics}
\label{sec:exp_setup}

\paragraph{PI-CAI: multimodal prostate imaging.}
PI-CAI~\citep{saha2024picai} pairs prostate MRI with patient information, including age, prostate-specific antigen (PSA), PSA density, and prostate volume, for clinically significant prostate cancer (csPCa) detection. We use institution (RUMC, ZGT, and PCNN) as the subgroup label. We hold out one fold of a patient-level five-fold split of the $1{,}500$ public cases, stratified by institution and csPCa status, as the test set ($301$ cases: RUMC $160$, ZGT $70$, PCNN $71$). We split the remaining $1{,}199$ cases at the patient level into $960$ training and $239$ validation cases.

\paragraph{Radiotherapy: multimodal \& unimodal prostate target volume delineation.}
The RT data consist of pelvic CT scans and clinician-drawn clinical target  volume (CTV) contours from three institutions, SC, YI, and GN. In the multimodal protocol, institution selects the router and linked records provide patient context, including age, PSA, grade, T/N stage, treatment information, and a de-identified physician code; the held-out set contains $420$ cases (SC $189$, YI $112$, GN $119$). The unimodal protocol uses CT and T stage for routing without a linked clinical-memory stream; its held-out set contains $275$ cases (T1 $11$, T2 $129$, T3 $114$, T4 $21$).

\paragraph{Harvard-FairSeg: unimodal retinal imaging.}
Harvard-FairSeg contains $10{,}000$ scanning laser ophthalmoscopy (SLO) fundus images with optic-cup and neuroretinal-rim masks used in glaucoma assessment. Among its six annotated demographic attributes, we use race ($\{\mathrm{Asian},\mathrm{Black},\mathrm{White}\}$) as the subgroup label, motivated by the benchmark's reported racial segmentation disparities and elevated glaucoma risk for Black patients~\citep{tianfairseg}. We evaluate both targets on the official $2{,}000$-image test split.

\paragraph{Baseline methods.}
For multimodal PI-CAI and RT, we compare \textbf{DuetMoE} with 3D ResUNet, dMoE~\citep{oh2025distribution}, MoME~\citep{oh2024mixture}, PCSA-Seg~\citep{XUE2026115745}, and a ConTEXTualNet3D-based baseline adapted to a 3D residual U-Net backbone~\citep{huemann2026contextual}. For unimodal RT and FairSeg, dMoE isolates subgroup routing; MoE, FEBS, GDRO, and Prompt-GDRO provide further controls~\citep{shazeer2017outrageously,tianfairseg,sagawa2019distributionally,panaganti2026groupd}. FairSeg also includes published TransUNet, ADV, FEBS, and FairDiff results~\citep{chen2021transunet,madras2018learning,tianfairseg,li2024fairdiff}. Complete implementation details and dataset cohorts are detailed in Appendix~\ref{app:implementation_details}. 

\paragraph{Evaluation metrics.}
For PI-CAI, the PI score averages case-level AUROC and lesion-level average precision~\citep{saha2024picai}. For RT and Harvard-FairSeg, Dice measures overlap between predicted and reference masks. We also report equity-scaled scores (ES-PI and ES-Dice), which adjust the overall scores for differences in subgroup performance~\citep{tianfairseg}; metric definitions are in Appendix~\ref{app:fairness_metrics}. For intra-subgroup evaluation, we report sensitivity at $90\%$ specificity on PI-CAI, patient-level Dice distributions and 25th-percentile Dice on RT, and the lowest-$10\%$ mean Dice within each racial subgroup on FairSeg.

\paragraph{Statistical analysis.}
To account for small and imbalanced subgroups, we estimate $95\%$ confidence intervals (CIs) from $1{,}000$ within-subgroup bootstrap resamples for methods with case-level predictions, including reproduced baselines.

\subsection{PI-CAI: multimodal prostate imaging}
\label{sec:results_picai}

Table~\ref{tab:picai_multimodal} compares the institution-grouped multimodal models on the PI-CAI test split. For inter-subgroup performance, \textbf{DuetMoE} raises ES-PI from $0.453$ with dMoE to $0.500$ and improves the lowest-scoring institution, PCNN, from $0.423$ to $0.457$. It also has the highest overall PI score ($0.623$) and PI score in RUMC ($0.688$). For intra-subgroup detection reliability, \textbf{DuetMoE} increases sensitivity at $90\%$ specificity within RUMC from $0.553$ with MoME to $0.681$ ($+12.8$ percentage points). It also achieves the highest normalized partial AUROC over false-positive rates up to $0.2$ in RUMC ($0.564$) and ZGT ($0.670$), supporting stronger detection performance at high specificity within these institutions. More institution-specific results are provided in Tables~\ref{tab:picai_sens_spec90} and~\ref{tab:picai_pauc} in Appendix~\ref{picai-intra}.

\begin{table}[h!]
\centering
\caption{\textbf{PI-CAI.} Comparison on \textbf{multimodal} csPCa detection using institution as the subgroup attribute.}
\label{tab:picai_multimodal}
\resizebox{\linewidth}{!}{%
\begin{tabular}{lccccc}
\toprule
Method & \multicolumn{2}{c}{All ($n=301$)} & RUMC ($n=160$) & ZGT ($n=70$) & PCNN ($n=71$) \\
\cmidrule(lr){2-3}\cmidrule(lr){4-4}\cmidrule(lr){5-5}\cmidrule(lr){6-6}
 & ES-PI (CIs) & PI score (CIs) & PI score (CIs) & PI score (CIs) & PI score (CIs) \\
\midrule
3D ResUNet  & 0.466 (0.379--0.535) & 0.585 (0.505--0.674) & 0.618 (0.523--0.732) & \underline{0.667} (0.473--0.867) & 0.445 (0.269--0.626) \\
ConTEXTualNet3D-based & 0.421 (0.342--0.455) & 0.486 (0.411--0.563) & 0.533 (0.437--0.658) & 0.496 (0.357--0.677) & 0.390 (0.277--0.532) \\
PCSA-Seg  & 0.432 (0.366--0.484) & 0.556 (0.487--0.643) & \underline{0.670} (0.580--0.760) & 0.537 (0.348--0.746) & 0.401 (0.302--0.553) \\
dMoE & 0.453 (0.371--0.526) & 0.590 (0.511--0.682) & 0.647 (0.554--0.768) & \textbf{0.671} (0.472--0.870) & 0.423 (0.271--0.596) \\
MoME & \textbf{0.510} (0.414--0.564) & \underline{0.617} (0.542--0.692) & 0.669 (0.575--0.777) & 0.641 (0.454--0.827) & \textbf{0.484} (0.331--0.635) \\
\textbf{DuetMoE} & \underline{0.500} (0.408--0.551) & \textbf{0.623} (0.543--0.705) & \textbf{0.688} (0.588--0.787) & 0.637 (0.451--0.839) & \underline{0.457} (0.310--0.616) \\
\bottomrule
\end{tabular}%
}
\par\smallskip
{\raggedright\footnotesize
Bold and underlined point estimates indicate the highest and second-highest values in each column, respectively.\par}
\end{table}

\subsection{Radiotherapy: multimodal \& unimodal prostate target volume delineation}
\label{sec:results_rt}

\paragraph{Multimodal RT: institution.} Table~\ref{tab:rt_multimodal} shows improved inter-subgroup performance for \textbf{DuetMoE}, with the highest ES-Dice ($0.716$, versus $0.686$ for MoME) and overall Dice ($0.761$). The gain is particularly relevant at GN, the lowest-scoring institution for every method: its Dice rises to $0.723$, from $0.693$ with dMoE and $0.698$ with MoME. The gap between the highest- and lowest-scoring institutions narrows from $0.088$ for MoME to $0.061$ for \textbf{DuetMoE}. 

\begin{table}[h!]
\centering
\caption{\textbf{Radiotherapy.} Comparison on \textbf{multimodal} 3D radiotherapy target segmentation using institution as the subgroup attribute.}
\label{tab:rt_multimodal}
\resizebox{\linewidth}{!}{%
\begin{tabular}{lccccc}
\toprule
Method & \multicolumn{2}{c}{All ($n=420$)} & SC ($n=189$) & YI ($n=112$) & GN ($n=119$) \\
\cmidrule(lr){2-3}\cmidrule(lr){4-4}\cmidrule(lr){5-5}\cmidrule(lr){6-6}
 & ES-Dice (CIs) & Dice (CIs) & Dice (CIs) & Dice (CIs) & Dice (CIs) \\
\midrule
3D ResUNet & 0.594 (0.566--0.614) & 0.674 (0.657--0.691) & 0.726 (0.701--0.751) & 0.673 (0.645--0.697) & 0.593 (0.557--0.624) \\
ConTEXTualNet3D-based & 0.640 (0.612--0.659) & 0.729 (0.712--0.746) & 0.781 (0.759--0.801) & 0.730 (0.702--0.756) & 0.644 (0.607--0.680) \\
PCSA-Seg& 0.627 (0.593--0.654) & 0.682 (0.665--0.699) & 0.701 (0.674--0.729) & 0.702 (0.670--0.733) & 0.633 (0.596--0.666) \\
dMoE  & 0.680 (0.647--0.704) & 0.747 (0.731--0.762) & \underline{0.784} (0.762--0.804) & 0.740 (0.710--0.767) & 0.693 (0.656--0.724) \\
MoME  & \underline{0.686} (0.659--0.712) & \underline{0.757} (0.743--0.771) & \textbf{0.786} (0.765--0.804) & \textbf{0.772} (0.744--0.799) & \underline{0.698} (0.663--0.731) \\
\textbf{DuetMoE} & \textbf{0.716} (0.679--0.737) & \textbf{0.761} (0.745--0.776) & \underline{0.784} (0.761--0.804) & \underline{0.761} (0.731--0.787) & \textbf{0.723} (0.687--0.755) \\
\bottomrule
\end{tabular}%
}
\par\smallskip
{\raggedright\footnotesize
Bold and underlined point estimates indicate the highest and second-highest values in each column, respectively; the tied second-highest SC scores are both underlined.\par}
\end{table}
\FloatBarrier

\begin{figure}[htbp]
    \centering
    \includegraphics[width=\linewidth]{Radiotherapy_analysis.png}
    \caption{\textbf{Multimodal radiotherapy: intra-subgroup performance.}
\textbf{Left:} patient-level Dice distributions in GN; violin width
reflects estimated density, points represent patients, and red horizontal lines mark the 25th percentiles. \textbf{Right:} selected target delineations from SC, YI, and GN for methods; green masks show target volumes and red contours indicate ground-truth boundaries.}
    \label{fig:radiotherapy_analysis}
\end{figure}

Figure~\ref{fig:radiotherapy_analysis} (left) shows improved intra-subgroup performance in GN: \textbf{DuetMoE} raises 25th-percentile Dice from $0.614$ with dMoE to $0.692$ ($+7.8$ percentage points), improving the lower end of the patient distribution. Violin width reflects the estimated density of patient Dice; \textbf{DuetMoE}'s thinner low-Dice tail suggests a lower concentration of poor predictions. The selected contours (right) show \textbf{DuetMoE}'s more complete target coverage in SC and YI and reduced extension into surrounding tissue in GN.

\paragraph{Unimodal RT: tumor stage.} Table~\ref{tab:rt_tstage} reports the image-only RT comparison across T1--T4. \textbf{DuetMoE+} obtains the highest overall Dice ($0.665$) and ES-Dice ($0.530$). Compared with dMoE~\citep{oh2025distribution}, it improves mean Dice in T2, the lowest-scoring stage, from $0.586$ to $0.621$ ($+3.5$ percentage points). ES-Dice rises from $0.499$ to $0.530$, supporting improved inter-subgroup performance. Additional evaluation metrics show a similar pattern, see Appendix~\ref{app:rt_tstage_supplement} for details.

\begin{table*}[h!]
\centering
\caption{Comparison on \textbf{unimodal} 3D radiotherapy target segmentation using tumor stage as the subgroup attribute.}
\label{tab:rt_tstage}
\resizebox{\linewidth}{!}{%
\setlength{\tabcolsep}{3pt}%
\begin{tabular}{lcccccc}
\toprule
Method & \multicolumn{2}{c}{All ($n=275$)} & T1 ($n=11$) & T2 ($n=129$) & T3 ($n=114$) & T4 ($n=21$) \\
\cmidrule(lr){2-3} \cmidrule(lr){4-4} \cmidrule(lr){5-5} \cmidrule(lr){6-6} \cmidrule(lr){7-7}
& ES-Dice (95\% CI) & Dice (95\% CI) & Dice (95\% CI) & Dice (95\% CI) & Dice (95\% CI) & Dice (95\% CI) \\
\midrule
3D ResUNet & 0.487 (0.447--0.529) & 0.610 (0.589--0.630) & 0.492 (0.401--0.579) & 0.570 (0.539--0.602) & 0.659 (0.626--0.688) & 0.656 (0.586--0.720) \\
+ FEBS& 0.434 (0.406--0.467) & 0.586 (0.567--0.604) & 0.466 (0.387--0.544) & 0.522 (0.495--0.549) & 0.652 (0.621--0.677) & 0.688 (0.651--0.721) \\
+ MoE & 0.452 (0.415--0.492) & 0.608 (0.586--0.628) & 0.492 (0.395--0.590) & 0.543 (0.512--0.573) & 0.674 (0.640--0.704) & 0.708 (0.663--0.750) \\
+ GDRO & 0.454 (0.421--0.491) & 0.607 (0.586--0.627) & 0.481 (0.400--0.560) & 0.546 (0.516--0.576) & 0.671 (0.639--0.700) & 0.694 (0.637--0.745) \\
+ Prompt-GDRO & 0.476 (0.440--0.517) & 0.612 (0.592--0.631) & 0.492 (0.409--0.568) & 0.562 (0.532--0.593) & 0.669 (0.636--0.697) & 0.673 (0.616--0.726) \\
+ dMoE & \underline{0.499 (0.469--0.531)} & \underline{0.650 (0.628--0.671)} & \underline{0.718 (0.654--0.772)} & \underline{0.586 (0.556--0.618)} & \textbf{0.693 (0.657--0.723)} & \textbf{0.778 (0.747--0.803)} \\
+ \textbf{DuetMoE+} & \textbf{0.530 (0.499--0.569)} & \textbf{0.665 (0.643--0.683)} & \textbf{0.757 (0.687--0.812)} & \textbf{0.621 (0.590--0.654)} & \underline{0.688 (0.654--0.717)} & \underline{0.758 (0.729--0.784)} \\
\bottomrule
\end{tabular}}
\par\smallskip
{\raggedright\footnotesize
Bold marks the highest score, and underlining marks the second-highest score in each column.\par}
\end{table*}

\begin{figure}[htbp]
\centering
\includegraphics[width=\linewidth]{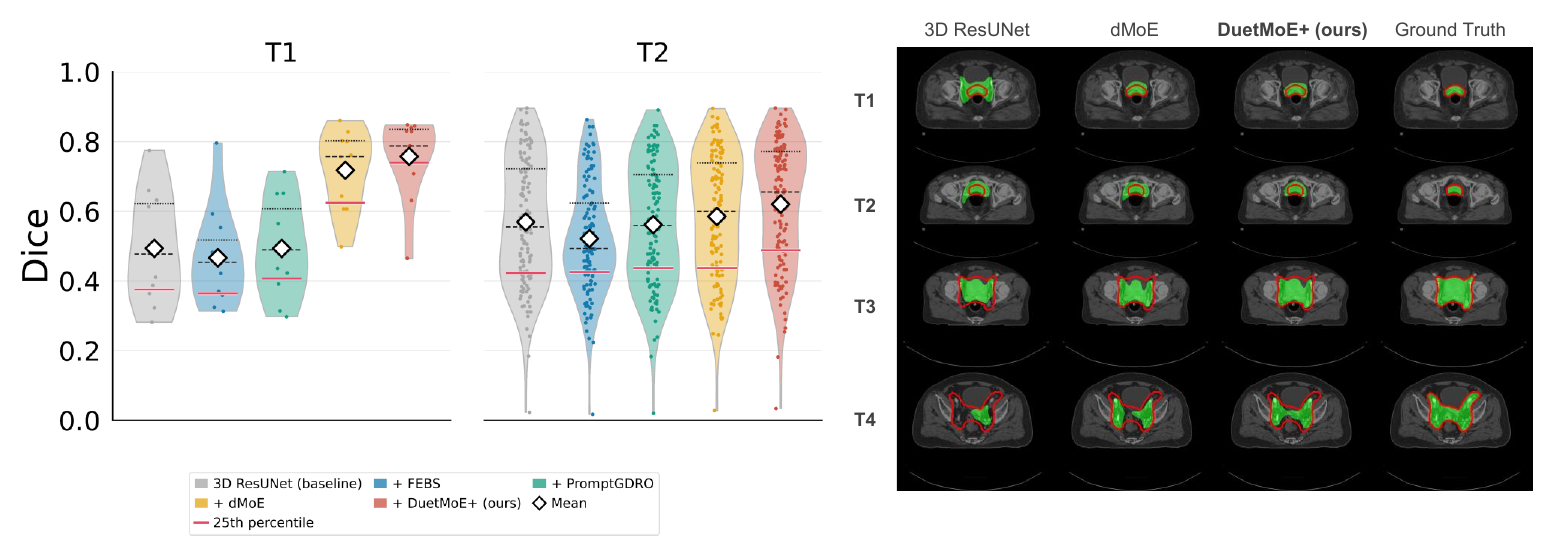}
\caption{\textbf{Unimodal radiotherapy: intra-subgroup performance.} \textbf{Left:} patient-level Dice distributions within T1 and T2. Violin width reflects estimated density, points represent patients, and red horizontal lines mark the 25th percentiles. \textbf{Right:} representative target delineations for T1--T4, comparing \textbf{DuetMoE+} with the baselines and ground truth. Green masks show target volumes, and red contours indicate ground-truth boundaries.}
\label{fig:rt_tstage_analysis}
\end{figure}

Figure~\ref{fig:rt_tstage_analysis} provides the complementary intra-subgroup view. Relative to dMoE, the T1 and T2 Dice distributions for \textbf{DuetMoE+} shift toward higher values, with 25th-percentile gains of $11.5$ percentage points ($0.625$ to $0.740$) and $5.1$ points ($0.437$ to $0.488$), respectively. In T2, the thinner lower tail also indicates less estimated density at low Dice. Since dMoE already uses subgroup-selected routing, these lower-tail gains support the contribution of robust training within each stage. The selected T1 and T2 contours show predictions that follow the annotated targets more closely.

\subsection{Harvard-FairSeg: unimodal retinal imaging}
\label{sec:results_fairseg}

Table~\ref{tab:fairseg-cup} reports cup-segmentation results on Harvard-FairSeg with race as the subgroup attribute. For inter-subgroup performance, \textbf{DuetMoE+} obtains the highest ES-Dice ($0.843$), up from $0.832$ with dMoE. It also raises mean Dice in the weakest subgroup, Asian, from $0.844$ to $0.852$ ($+0.8$ percentage points), with an overall Dice of $0.865$. For intra-subgroup performance, Figure~\ref{fig:fairseg_tail} shows that \textbf{DuetMoE+} increases mean cup Dice among the lowest-scoring $10\%$ of patients by $6.5$, $8.4$, and $1.4$ percentage points over dMoE in the Asian, Black, and White subgroups, respectively. Rim segmentation also shows lower-tail gains in the Asian and Black subgroups; more results and additional evaluation metrics are reported in Appendix~\ref{app:fairseg-rim}.

\begin{table*}[h!]
\centering
\caption{Cup Segmentation Comparison on the \textbf{unimodal} 2D Harvard-FairSeg dataset using race as the subgroup attribute.}
\label{tab:fairseg-cup}
\resizebox{\linewidth}{!}{%
\begin{tabular}{lccccc}
\toprule
Method & \multicolumn{2}{c}{All ($n=2000$)} & Asian ($n=169$) & Black ($n=299$) & White ($n=1532$) \\
\cmidrule(lr){2-3} \cmidrule(lr){4-4} \cmidrule(lr){5-5} \cmidrule(l){6-6}
 & ES-Dice (CIs) & Dice (CIs) & Dice & Dice & Dice \\
\midrule
\cmidrule(l){1-6}
TransUNet & 0.828 & 0.848 & 0.827 & 0.849 & 0.850 \\
+ ADV & 0.826 & 0.841 & 0.825 & 0.842 & 0.843 \\
+ FEBS & 0.825 & 0.846 & 0.825 & 0.848 & 0.848 \\
+ FairDiff & 0.833 & 0.848 & 0.832 & 0.848 & 0.850 \\
+ MoE & 0.830 (0.809--0.847) & 0.854 (0.849--0.860) & 0.845 & 0.842 & 0.857 \\
+ GDRO & 0.840 (0.818--0.857) & 0.860 (0.855--0.865) & \underline{0.850} & 0.851 & 0.863 \\
+ Prompt-GDRO & \underline{0.842} (0.819--0.861) & \textbf{0.866} (0.861--0.871) & 0.849 & \textbf{0.859} & \textbf{0.870} \\
+ dMoE & 0.832 (0.810--0.853) & 0.862 (0.856--0.867) & 0.844 & 0.851 & 0.866 \\
+ \textbf{DuetMoE+} & \textbf{0.843} (0.822--0.861) & \underline{0.865} (0.860--0.869) & \textbf{0.852} & \underline{0.856} & \underline{0.868} \\
\bottomrule
\end{tabular}}
\par\smallskip
{\raggedright\footnotesize
Bold and underlined point estimates indicate the highest and second-highest values within each target and column, respectively.\par}
\end{table*}

\section{Discussion \& Conclusion}
\label{sec:discussion_conclusion}

Our results support treating inter-subgroup balance and intra-subgroup robustness as complementary goals for medical image analysis. Across four evaluation settings, \textbf{DuetMoE} and \textbf{DuetMoE+} achieve leading overall or equity-scaled performance. On PI-CAI, \textbf{DuetMoE} achieves the highest overall PI score ($0.623$) and improves sensitivity at $90\%$ specificity within RUMC by $10.7$ percentage points over the strongest baseline for this metric. In multimodal RT, \textbf{DuetMoE} narrows the inter-institution mean group Dice gap by $30.7\%$ relative to the best prior model in overall Dice. For unimodal setting, in image-only RT, \textbf{DuetMoE+} improves worst-subgroup mean Dice by $3.5$ percentage points and within-subgroup 25th-percentile Dice by up to $11.5$ points over the subgroup-routing baseline. On Harvard-FairSeg, it also improves lowest-$10\%$ mean cup Dice by up to $8.4$ percentage points over the same baseline. These within-subgroup gains demonstrate the value of evaluating patients within each subgroup alongside aggregate performance.

These results support embedding patient-level adaptation within subgroup-aware learning. \textbf{DuetMoE} embeds clinical memory before and after routing to guide expert selection and feature processing. Without linked clinical records, \textbf{DuetMoE+} aligns within-subgroup robust weighting with the subgroup routers and shared experts that process each patient's images. The PI-CAI clinical-context comparisons favor designs with expert-specific guidance (Appendix~\ref{app:duetmoe_context_configurations}), while the unimodal component ablation shows that the full \textbf{DuetMoE+} achieves the highest equity-scaled and worst-subgroup scores (Appendix~\ref{app:ablation_two_axes}). The radius study in Appendix~\ref{app:rho_selection} further shows that settings with similar overall Dice can yield different subgroup lower-tail outcomes. For \textbf{DuetMoE+}, radius selection should therefore assess equity-scaled performance and subgroup lower tails alongside overall Dice on a held-out validation set.

In particular, both methods require subgroup labels at inference, and \textbf{DuetMoE} additionally relies on linked clinical records. Their sensitivity to missing or inaccurate subgroup metadata and incomplete clinical context remains to be evaluated. These observations motivate extending the framework to incomplete information and developing validation-based radius selection that considers both subgroup performance and patient-level lower tails. A further direction is to combine loss-level robustness training with clinically guided \textbf{DuetMoE}, allowing patient-specific computation and loss-based emphasis to act together. Since both mechanisms influence learning within a subgroup, their interaction could either reinforce useful adaptation or place excessive emphasis on noisy or persistently difficult cases. A matched multimodal ablation is needed to test whether the two mechanisms complement or conflict with each other. Overall, our current findings support models that adapt across subgroups while remaining responsive to patient differences within each group.

\section*{Acknowledgments}
We thank the clinical teams at the three participating hospitals and the teams behind PI-CAI and Harvard-FairSeg for their contributions to this study. The retrospective radiotherapy cohort was collected under institutional review board approvals at SC (4-2023-0179), YI (9-2023-0161), and GN (3-2023-0396). Patient records were de-identified, and access to the in-house cohort remains subject to institutional and IRB restrictions.

\clearpage
\bibliography{duetfair_drafts_refs}

@inproceedings{tian2024fairdomain,
  title={Fairdomain: Achieving fairness in cross-domain medical image segmentation and classification},
  author={Tian, Yu and Wen, Congcong and Shi, Min and Afzal, Muhammad Muneeb and Huang, Hao and Khan, Muhammad Osama and Luo, Yan and Fang, Yi and Wang, Mengyu},
  booktitle={European Conference on Computer Vision},
  pages={251--271},
  year={2024},
  organization={Springer},
  doi={10.1007/978-3-031-73116-7_15}
}

@article{chen2021transunet,
  title={Transunet: Transformers make strong encoders for medical image segmentation},
  author={Chen, Jieneng and Lu, Yongyi and Yu, Qihang and Luo, Xiangde and Adeli, Ehsan and Wang, Yan and Lu, Le and Yuille, Alan L and Zhou, Yuyin},
  journal={arXiv preprint arXiv:2102.04306},
  year={2021}
}

@inproceedings{madras2018learning,
  title={Learning adversarially fair and transferable representations},
  author={Madras, David and Creager, Elliot and Pitassi, Toniann and Zemel, Richard},
  booktitle={International Conference on Machine Learning},
  pages={3384--3393},
  year={2018},
  organization={PMLR}
}

@article{sagawa2019distributionally,
  title={Distributionally robust neural networks for group shifts: On the importance of regularization for worst-case generalization},
  author={Sagawa, Shiori and Koh, Pang Wei and Hashimoto, Tatsunori B and Liang, Percy},
  journal={arXiv preprint arXiv:1911.08731},
  year={2019},
  doi={10.48550/arXiv.1911.08731}
}

@inproceedings{tianfairseg,
  title={FairSeg: A Large-Scale Medical Image Segmentation Dataset for Fairness Learning Using Segment Anything Model with Fair Error-Bound Scaling},
  author={Tian, Yu and Shi, Min and Luo, Yan and Kouhana, Ava and Elze, Tobias and Wang, Mengyu},
  booktitle={The Twelfth International Conference on Learning Representations},
  year={2024},
  doi={10.48550/arXiv.2311.02189}  
}

@article{oh2023llm,
  title={LLM-driven multimodal target volume contouring in radiation oncology},
  author={Oh, Yujin and Park, Sangjoon and Byun, Hwa Kyung and Cho, Yeona and Lee, Ik Jae and Kim, Jin Sung and Ye, Jong Chul},
  journal={Nature Communications},
  volume={15},
  number={1},
  pages={9186},
  year={2024},
  publisher={Nature Publishing Group UK London},
  doi={10.1038/s41467-024-53387-y}
}

@inproceedings{cciccek20163d,
  title={{3D U-Net: learning dense volumetric segmentation from sparse annotation}},
  author={{\c{C}}i{\c{c}}ek, {\"O}zg{\"u}n and Abdulkadir, Ahmed and Lienkamp, Soeren S and Brox, Thomas and Ronneberger, Olaf},
  booktitle={Medical Image Computing and Computer-Assisted Intervention},
  pages={424--432},
  year={2016},
  organization={Springer}
}

@inproceedings{li2024fairdiff,
  title={Fairdiff: Fair segmentation with point-image diffusion},
  author={Li, Wenyi and Xu, Haoran and Zhang, Guiyu and Gao, Huan-ang and Gao, Mingju and Wang, Mengyu and Zhao, Hao},
  booktitle={International Conference on Medical Image Computing and Computer-Assisted Intervention},
  pages={617--628},
  year={2024},
  organization={Springer},
  doi={10.1007/978-3-031-72384-1_58}
}

@article{choi2020clinical,
  title={Clinical evaluation of atlas- and deep learning-based automatic segmentation of multiple organs and clinical target volumes for breast cancer},
  author={Choi, Min Seo and Choi, Byeong Su and Chung, Seung Yeun and Kim, Nalee and Chun, Jaehee and Kim, Yong Bae and Chang, Jee Suk and Kim, Jin Sung},
  journal={Radiotherapy and Oncology},
  volume={153},
  pages={139--145},
  year={2020},
  publisher={Elsevier}
}

@article{paszke2019pytorch,
  title={Pytorch: An imperative style, high-performance deep learning library},
  author={Paszke, Adam and Gross, Sam and Massa, Francisco and Lerer, Adam and Bradbury, James and Chanan, Gregory and Killeen, Trevor and Lin, Zeming and Gimelshein, Natalia and Antiga, Luca and others},
  journal={Advances in neural information processing systems},
  volume={32},
  year={2019}
}

@article{shazeer2017outrageously,
  title={Outrageously large neural networks: The sparsely-gated mixture-of-experts layer},
  author={Shazeer, Noam and Mirhoseini, Azalia and Maziarz, Krzysztof and Davis, Andy and Le, Quoc and Hinton, Geoffrey and Dean, Jeff},
  journal={arXiv preprint arXiv:1701.06538},
  year={2017}
}

@article{oh2024mixture,
  title={Mixture of Multicenter Experts in Multimodal AI for Debiased Radiotherapy Target Delineation},
  author={Oh, Yujin and Park, Sangjoon and Li, Xiang and Jin, Pengfei and Wang, Yi and Paly, Jonathan and Efstathiou, Jason and Chan, Annie and Kim, Jun Won and Byun, Hwa Kyung and Lee, Ik Jae and Cho, Jaeho and Wee, Chan Woo and Shu, Peng and Wang, Peilong and Jiang, Caiwen and Yu, Nathan and Holmes, Jason and Ye, Jong Chul and Li, Quanzheng and Liu, Wei and Koom, Woong Sub and Kim, Jin Sung and Kim, Kyungsang},
  journal={arXiv preprint arXiv:2410.00046},
  year={2024},
  doi={10.48550/arXiv.2410.00046}
}

@article{curi2020adaptive,
  title={Adaptive sampling for stochastic risk-averse learning},
  author={Curi, Sebastian and Levy, Kfir Y and Jegelka, Stefanie and Krause, Andreas},
  journal={Advances in Neural Information Processing Systems},
  volume={33},
  pages={1036--1047},
  year={2020},
  doi={10.48550/arXiv.1910.12511}  
}

@inproceedings{oh2025distribution,
  title={Distribution-aware Fairness Learning in Medical Image Segmentation From A Control-Theoretic Perspective},
  author={Oh, Yujin and Jin, Pengfei and Park, Sangjoon and Kim, Sekeun and Yoon, Siyeop and Kim, Jin Sung and Kim, Kyungsang and Li, Xiang and Li, Quanzheng},
  booktitle={Proceedings of the 42nd International Conference on Machine Learning},
  series={Proceedings of Machine Learning Research},
  volume={267},
  pages={47003--47016},
  year={2025},
  publisher={PMLR},
  url={https://proceedings.mlr.press/v267/oh25c.html},
  doi={10.48550/arXiv.2502.00619}  
}

@article{loshchilov2017decoupled,
  title={Decoupled weight decay regularization},
  author={Loshchilov, Ilya and Hutter, Frank},
  journal={arXiv preprint arXiv:1711.05101},
  year={2017}
}

@misc{panaganti2026groupd,
      title={Group Distributionally Robust Optimization-Driven Reinforcement Learning for LLM Reasoning}, 
      author={Kishan Panaganti and Zhenwen Liang and Wenhao Yu and Haitao Mi and Dong Yu},
      year={2026},
      eprint={2601.19280},
      archivePrefix={arXiv},
      primaryClass={cs.LG},
      url={https://arxiv.org/abs/2601.19280}
}

@article{liu2023translational,
  title={A translational perspective towards clinical AI fairness},
  author={Liu, Mingxuan and Ning, Yilin and Teixayavong, Salinelat and Mertens, Mayli and Xu, Jie and Ting, Daniel Shu Wei and Cheng, Lionel Tim-Ee and Ong, Jasmine Chiat Ling and Teo, Zhen Ling and Tan, Ting Fang and others},
  journal={NPJ digital medicine},
  volume={6},
  number={1},
  pages={172},
  year={2023},
  publisher={Nature Publishing Group UK London}
}

@article{saha2024picai,
  title={Artificial intelligence and radiologists in prostate cancer detection on MRI (PI-CAI): an international, paired, non-inferiority, confirmatory study},
  author={Saha, Anindo and Bosma, Joeran S. and Twilt, Jasper J. and van Ginneken, Bram and Bjartell, Anders and Padhani, Anwar R. and others},
  journal={The Lancet Oncology},
  volume={25},
  number={7},
  pages={879--887},
  year={2024},
  doi={10.1016/S1470-2045(24)00220-1}
}

@inproceedings{oakdenrayner2020hidden,
  title={Hidden Stratification Causes Clinically Meaningful Failures in Machine Learning for Medical Imaging},
  author={Oakden-Rayner, Luke and Dunnmon, Jared and Carneiro, Gustavo and Re, Christopher},
  booktitle={Proceedings of the ACM Conference on Health, Inference, and Learning},
  year={2020},
  doi={10.1145/3368555.3384468}
}

@article{rajendran2025autodelineation,
  title={Auto-delineation of Treatment Target Volume for Radiation Therapy Using Large Language Model-Aided Multimodal Learning},
  author={Rajendran, Praveenbalaji and Chen, Yizheng and Qiu, Liang and Niedermayr, Thomas and Liu, Wu and Buyyounouski, Mark and Bagshaw, Hilary and Han, Bin and Yang, Yong and Kovalchuk, Nataliya and Gu, Xuejun and Hancock, Steven and Xing, Lei and Dai, Xianjin},
  journal={International Journal of Radiation Oncology, Biology, Physics},
  year={2025},
  doi={10.1016/j.ijrobp.2024.07.2149}
}

@inproceedings{tang2025ltse,
  title={{LTSE}: Language-guided Tissue Referring Segmentation in Pathology Images with Adaptive Expert Mixture},
  author={Tang, Jiao and Qian, Bo and Wan, Peng and Shao, Wei and Zhang, Daoqiang},
  booktitle={Medical Image Computing and Computer Assisted Intervention -- MICCAI 2025},
  series={Lecture Notes in Computer Science},
  volume={15965},
  pages={405--415},
  year={2025},
  doi={10.1007/978-3-032-04978-0_39}
}

@article{li2023tilted,
  title={On Tilted Losses in Machine Learning: Theory and Applications},
  author={Li, Tian and Beirami, Ahmad and Sanjabi, Maziar and Smith, Virginia},
  journal={Journal of Machine Learning Research},
  volume={24},
  number={142},
  pages={1--79},
  year={2023},
  url={https://jmlr.org/papers/v24/21-1095.html},
  doi={10.48550/arXiv.2109.06141}  
}

@inproceedings{jo2026hierarchical,
 author = {Jo, Sungho and Kim, Seonghwi and Chae, Minwoo},
 booktitle = {International Conference on Learning Representations},
 editor = {C. Vondrick and B. Hariharan and C. Raffel and L. Pinto and D. Yang and A. Faust},
 pages = {121167--121191},
 title = {Mitigating Spurious Correlation via Distributionally Robust Learning with Hierarchical Ambiguity Sets},
 url = {https://proceedings.iclr.cc/paper_files/paper/2026/file/c4f85da01632a2211c785f482cb3e043-Paper-Conference.pdf},
 volume = {2026},
 year = {2026}
}

@article{xiao2026medicalmllm,
  title={Medical multimodal large language models: A survey},
  author={Xiao, Hanguang and Hui, Ningzhi and Xu, Yong and Li, Zhipeng and Peng, Jincheng},
  journal={Information Fusion},
  volume={134},
  pages={104386},
  year={2026},
  doi={10.1016/j.inffus.2026.104386}
}

@article{XUE2026115745,
  title={Progressive cross-scale semantic alignment for language-guided medical image segmentation},
  author={Xue, Hengzhi and Dai, Yin and Li, Qingyong and Yao, Yudong and Teng, Yueyang},
  journal={Knowledge-Based Systems},
  volume={340},
  pages={115745},
  year={2026},
  doi={10.1016/j.knosys.2026.115745}
}

@inproceedings{lester2021power,
  title={The Power of Scale for Parameter-Efficient Prompt Tuning},
  author={Lester, Brian and Al-Rfou, Rami and Constant, Noah},
  booktitle={Proceedings of the 2021 Conference on Empirical Methods in Natural Language Processing},
  pages={3045--3059},
  year={2021},
  publisher={Association for Computational Linguistics},
  doi={10.18653/v1/2021.emnlp-main.243}
}

@article{huemann2026contextual,
  title={ConTEXTual Net 3D: Vision-Language Modeling in PET/CT for Visual Grounding of Positive Findings},
  author={Huemann, Zachary and Church, Samuel and Warner, Joshua D and Tran, Daniel and Tie, Xin and McMillan, Alan B and Hu, Junjie and Cho, Steve Y and Lubner, Meghan G and Bradshaw, Tyler J},
  journal={Journal of Imaging Informatics in Medicine},  
  year={2026},  
  publisher={Springer},
  doi={10.1007/s10278-026-01879-2} 
}
\bibliographystyle{plainnat}
\clearpage
\appendix
\section{Additional Methodological Details}
\subsection{DuetMoE: Architecture, Clinical Memory, and Computation}
\label{app:duetmoe_details}
The following diagrams show the DuetMoE block and the segmentation architecture.

\begin{figure}[htbp]
    \centering
    \includegraphics[width=0.8\linewidth]{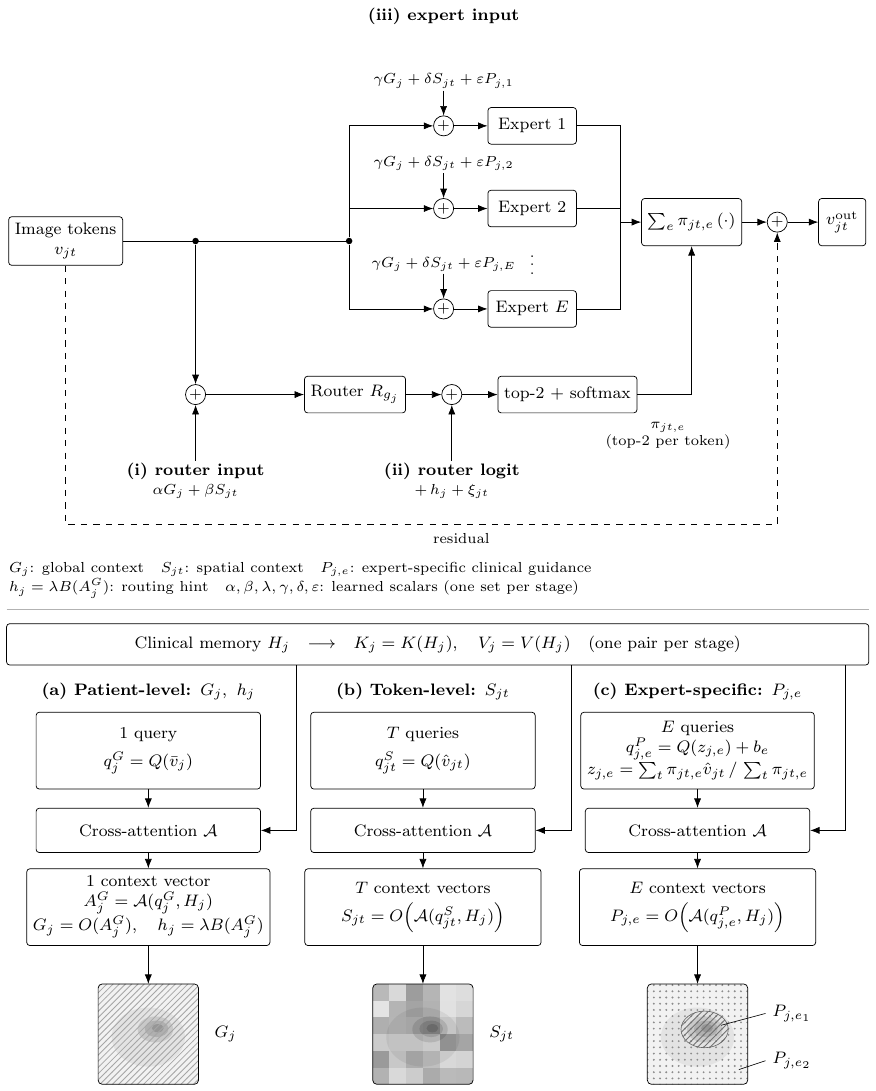}
    \caption{DuetMoE block and clinical-memory reads. Expert-specific queries are evaluated only for patient--expert pairs with positive routing mass, giving at most $E$ queries per patient and stage.}
    \label{fig:duetmoe_flow}
\end{figure}

\begin{figure}[htbp]
    \centering
    \includegraphics[width=0.8\linewidth]{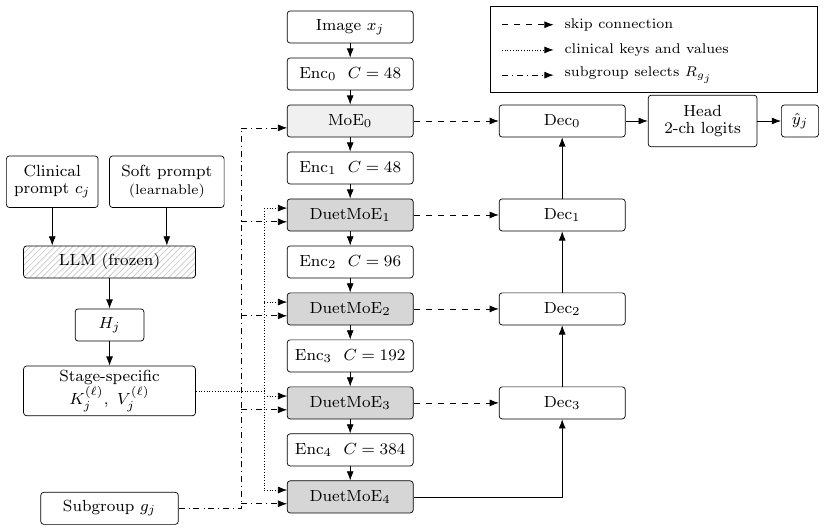}
    \caption{DuetMoE segmentation architecture.}
    \label{fig:duetmoe_encoder}
\end{figure}

\paragraph{Clinical-memory construction.}
In the PI-CAI implementation, the clinical record $c_j$ is serialized as a prompt with tags delimiting its fields. A shared learnable soft-prompt sequence is inserted after the beginning-of-sequence token, and a frozen language model encodes the resulting sequence. Hidden states from the tagged conditioning spans form the memory vectors $h_{jm}$. For numerical fields, an encoding of the value and an observation indicator is appended to the corresponding tagged representation. Padding is masked, so $M_j$ counts only valid memory vectors. The same $H_j$ supplies keys and values to all clinical reads for patient $j$.

\paragraph{Attention and routing.}
At each clinical-context stage, $\operatorname{LN}$ normalizes each image token over its channels. The global query uses the mean $\bar v_j$, while each spatial query uses its own $\hat v_{jt}$. Stage-specific linear projections $Q$ and $O$ and the attention operator $\mathcal A$ produce Eq.~\eqref{eq:duetmoe_context}. The global attention output before $O$ supplies the expert-wise hint $B(A_j^G)$. For the selected set $\mathcal K_{jt}$ of $K$ experts, the sparse weight in Eq.~\eqref{eq:duetmoe_router} is
\begin{equation}
{\pi_{jt,e}=\begin{cases}
\displaystyle\frac{\exp(r_{jt,e})}{\sum_{k\in\mathcal K_{jt}}\exp(r_{jt,k})},&e\in\mathcal K_{jt},\\[4pt]
0,&e\notin\mathcal K_{jt}.
\end{cases}}
\end{equation}
The noise $\xi_{jt}$ perturbs scores during both training and evaluation, with a fixed random seed at evaluation. The subgroup label selects $R_{g_j}$, while the experts are shared across subgroups.

\paragraph{Coefficients and initialization.}
The router-input coefficients satisfy $\alpha,\beta\in[0,1]$, the hint coefficient $\lambda\geq0$, and the expert-input coefficients $\gamma,\delta,\varepsilon\geq0$ with $\gamma+\delta+\varepsilon\leq1$; the constraints are enforced by projection after every update. They are initialized at $\alpha=\beta=\lambda=1/7$ and $\gamma=\delta=\varepsilon=2/7$, held fixed for the first ten epochs, and then learned per stage. The output projection $O$ and the hint projection $B$ are initialized to zero, so every block starts as an image-only routed MoE layer and the clinical reads enter gradually during training.

\paragraph{Blocks, experts, and memory.}
One MoE block follows each of the five encoder stages; the first block has no clinical reads, and the decoder contains no MoE blocks. Each block has $E=8$ experts with top-$2$ routing, one router per subgroup, and experts shared across subgroups. An expert is a two-layer MLP with hidden width $4C$, ReLU, and dropout $0.1$, where $C$ is the stage width; the routing noise $\xi_{jt}$ is Gaussian with a learned, softplus-scaled amplitude~\citep{shazeer2017outrageously}. The attention operator $\mathcal A$ uses four heads, and each stage has one pair of key and value projections of memory vectors, shared by the global, spatial, and expert-specific reads.

\paragraph{Expert-specific read and update.}
For each patient--expert pair with positive routing mass, Eq.~\eqref{eq:duetmoe_guidance_main} averages the normalized tokens $\hat v_{jt}$ using their routing weights $\pi_{jt,e}$. The resulting summary $z_{j,e}$, together with the expert-specific offset $b_e$, queries the same patient memory to obtain $P_{j,e}$. Expert pairs with no selected tokens make no contribution. Equation~\eqref{eq:duetmoe_output} then adds the weighted expert outputs to the original token. All six clinical coefficients are learned per stage, and only the last four of the five encoder stages use these reads.

\paragraph{Language model.}
The clinical memory is produced by Meta-Llama-3.1-8B-Instruct, kept frozen, with $32$ learnable soft-prompt tokens. In radiotherapy, the encoded sequence has $96$ positions (the beginning-of-sequence token, the $32$ soft-prompt tokens, and $63$ report tokens), and all of them are retained as memory vectors of width $4{,}096$. In PI-CAI, the tokenized record has at most $384$ positions before insertion of the $32$ soft-prompt tokens; only the tagged conditioning spans are retained in memory. There, each memory vector carries an $11$-dimensional numeric tail that can be nonzero only at the first position of a numeric field's tagged span (width $4{,}107$).

\subsection{DuetMoE+: reweighting, KL reformulation, and sensitivity to intra-subgroup variation}
\label{app:duetmoeplus_details}

We prove the architectural reduction in Proposition~\ref{prop:duetmoe_context_free}, derive the finite-sample representation of the within-subgroup robust risk in Eq.~\eqref{eq:duetmoeplus_group_risk}, and prove Proposition~\ref{prop:duetmoeplus_variance}. Throughout the robust-risk analysis, $I_g=\{j:g_j=g\}$ is a nonempty subgroup, $n_g=|I_g|$, and $L_j(\theta)$ is the finite loss of patient $j$. The parameter vector $\theta$ includes all trainable model parameters. The nominal distribution is uniform over the observed patients in each subgroup, and all logarithms are natural.

\subsubsection{Proof of Proposition~\ref{prop:duetmoe_context_free}}
\label{app:duetmoe_context_free_proof}

\begin{proof}
Fix an encoder stage, its input tokens, subgroup routers, shared experts, and gating rule. Disabling the clinical pathways removes their additive contributions in Eqs.~\eqref{eq:duetmoe_router} and~\eqref{eq:duetmoe_output}. Algebraically, this is achieved by setting the six clinical coefficients $\alpha,\beta,\lambda,\gamma,\delta,\varepsilon$ to zero. The router and expert inputs consequently become
\begin{equation}
u_{jt}=v_{jt},\qquad
r_{jt}=R_{g_j}(v_{jt})+\xi_{jt},\qquad
\widetilde v_{jt,e}=v_{jt}.
\label{eq:duetmoe_context_free_inputs}
\end{equation}
Thus, with the same top-$K$ selection and tie-breaking rule,
\begin{equation}
\begin{aligned}
\pi_{jt}^{\mathrm{dMoE}}
&=\operatorname{SoftmaxTopK}\!\left(R_{g_j}(v_{jt})+\xi_{jt}\right),\\
v_{jt}^{\mathrm{out}}
&=v_{jt}+\sum_{e\in\mathcal K_{jt}^{\mathrm{dMoE}}}
\pi_{jt,e}^{\mathrm{dMoE}}E_e(v_{jt}),
\end{aligned}
\label{eq:duetmoe_context_free_update}
\end{equation}
where $\mathcal K_{jt}^{\mathrm{dMoE}}$ is the selected expert set. This is the subgroup-routed dMoE residual update with the stipulated routers, experts, and gating rule. It holds for every patient and image token. When gating or expert computation is stochastic, equality holds for the same realization of the random draws in both architectures; under the same randomization law, the updates therefore also have the same distribution. No clinical-memory read remains in the update, proving the proposition.
\end{proof}

This is an architectural identity for fixed model components. It does not assert that models trained with different loss objectives attain the same parameters.

\subsubsection{Two-level reweighting and composite sample weights}
\label{app:duetmoeplus_two_level}

Consider a weighted empirical objective with subgroup weights $\alpha_g\geq0$, $\sum_g\alpha_g=1$, and within-subgroup patient weights $\beta_j^{(g)}\geq0$, $\sum_{j\in I_g}\beta_j^{(g)}=1$:
\begin{equation}
\min_\theta\sum_{g\in\mathcal G}\alpha_g
\sum_{j\in I_g}\beta_j^{(g)}L_j(\theta).
\label{eq:duetmoeplus_two_level_loss}
\end{equation}
Here, $\alpha_g$ controls the total mass assigned to subgroup $g$, whereas $\beta_j^{(g)}$ distributes that mass among its patients. For fixed weights and differentiable patient losses,
\begin{equation}
\nabla_\theta\!\left(\sum_{g\in\mathcal G}\alpha_g
\sum_{j\in I_g}\beta_j^{(g)}L_j(\theta)\right)
=\sum_{g\in\mathcal G}\sum_{j\in I_g}
\alpha_g\beta_j^{(g)}\nabla_\theta L_j(\theta).
\label{eq:duetmoeplus_composite_gradient}
\end{equation}
Thus, the coefficient of a patient's loss gradient is the product $\alpha_g\beta_j^{(g)}$. An individual coefficient can reflect a larger subgroup mass, a larger within-subgroup weight, or both. This is a statement about an individual coefficient: given the subgroup partition and the complete normalized weight vector $\omega_j=\alpha_g\beta_j^{(g)}$, one can recover $\alpha_g=\sum_{j\in I_g}\omega_j$ and, when $\alpha_g>0$, $\beta_j^{(g)}=\omega_j/\alpha_g$.

In DuetMoE+, subgroup labels also select the router that processes the image features. This architectural role is distinct from assigning subgroup mass in a loss aggregation. The objective in Eq.~\eqref{eq:duetmoeplus_group_risk} uses a sum of subgroup risks, with each subgroup coefficient fixed at one; its inner KL optimization changes patient weights within each subgroup. Averaging this sum over $|\mathcal G|$ would multiply the objective and its gradient by $1/|\mathcal G|$. The fixed-weight calculation above is therefore a comparison of loss coefficients, rather than an additional outer adversarial optimization in DuetMoE+.

\subsubsection{KL reformulation and patient-weight interpretation}
\label{app:duetmoeplus_kl_reformulation}

Let $u_g=(1/n_g,\ldots,1/n_g)$ and
\begin{equation}
\Delta_{n_g}=\left\{q\in\mathbb R_+^{n_g}:
\sum_{j\in I_g}q_j=1\right\},\qquad
D_{\mathrm{KL}}(q\Vert u_g)
=\sum_{j\in I_g}q_j\log(n_gq_j),
\label{eq:conditional_kl_simplex}
\end{equation}
with $0\log0=0$. A distribution on the observed patients is represented by $q\in\Delta_{n_g}$. The robust subgroup risk is
\begin{equation}
\widehat R_g^{\mathrm{rob}}(\theta)
=\max_{q\in\Delta_{n_g}}
\left\{\sum_{j\in I_g}q_jL_j(\theta):
D_{\mathrm{KL}}(q\Vert u_g)\leq\rho_g\right\}.
\label{eq:conditional_kl_primal}
\end{equation}
The feasible set is nonempty and compact, since it contains $u_g$ and KL divergence is continuous on the finite simplex under the convention above. The maximum is therefore attained. This formulation keeps the observed patients fixed and varies their probabilities within a KL neighborhood of the uniform empirical distribution.

For every $\rho_g\geq0$, the exact dual representation is
\begin{equation}
\widehat R_g^{\mathrm{rob}}(\theta)
=\inf_{\eta_g>0}\left\{
\eta_g\rho_g+\eta_g\log\!\left(
\frac{1}{n_g}\sum_{j\in I_g}
\exp\!\left(\frac{L_j(\theta)}{\eta_g}\right)
\right)\right\}.
\label{eq:conditional_kl_dual}
\end{equation}
The use of an infimum includes boundary cases in which no finite positive dual optimizer exists. For $\eta_g>0$, define
\begin{equation}
q_j^{(g)}(\eta_g)
=\frac{\exp(L_j(\theta)/\eta_g)}
{\sum_{k\in I_g}\exp(L_k(\theta)/\eta_g)},
\qquad j\in I_g.
\label{eq:conditional_kl_weights}
\end{equation}

\paragraph{Derivation and boundary cases.}
Fix $g$ and $\theta$, and temporarily suppress the subgroup index. Write
$Z(\eta)=n^{-1}\sum_j\exp(L_j/\eta)$ and let $q(\eta)$ be the weights in Eq.~\eqref{eq:conditional_kl_weights}. Direct substitution gives, for every $q\in\Delta_n$,
\begin{equation}
\sum_jq_jL_j-\eta D_{\mathrm{KL}}(q\Vert u)
=\eta\log Z(\eta)-\eta D_{\mathrm{KL}}(q\Vert q(\eta)).
\label{eq:conditional_kl_penalty_identity}
\end{equation}
Since KL divergence is nonnegative, maximizing the left-hand side over the simplex gives $\eta\log Z(\eta)$, uniquely attained at $q(\eta)$. Every feasible $q$ consequently satisfies
$\sum_jq_jL_j\leq\eta\rho+\eta\log Z(\eta)$, proving the upper bound in Eq.~\eqref{eq:conditional_kl_dual}.

To establish equality and characterize attainment, define
\begin{equation}
K(t)=\log\!\left(\frac{1}{n}\sum_j e^{tL_j}\right),
\qquad p_j(t)=\frac{e^{tL_j}}{\sum_k e^{tL_k}},
\qquad d(t)=tK'(t)-K(t),\quad t\geq0.
\label{eq:conditional_kl_tilt}
\end{equation}
These quantities satisfy $d(t)=D_{\mathrm{KL}}(p(t)\Vert u)$ and
\begin{equation}
K'(t)=\sum_jp_j(t)L_j,\qquad
K''(t)=\sum_jp_j(t)\bigl(L_j-K'(t)\bigr)^2,
\qquad d'(t)=tK''(t).
\label{eq:conditional_kl_tilt_derivatives}
\end{equation}
Let $M=\{j:L_j=\max_k L_k\}$, $m=|M|$, and
$\rho_{\mathrm{sat}}=\log(n/m)$. If the losses are not all equal, $K''(t)>0$ for finite $t$, because every $p_j(t)$ is positive. Thus $d$ is strictly increasing on $(0,\infty)$, starts at $d(0)=0$, and converges to $\rho_{\mathrm{sat}}$: the weights converge to the uniform distribution on $M$, whose divergence from $u$ is $\log(n/m)$.

For $0<\rho<\rho_{\mathrm{sat}}$, there is a unique $t_\rho>0$ satisfying $d(t_\rho)=\rho$. Set $\eta^*=1/t_\rho$. Then $p(t_\rho)=q(\eta^*)$ is feasible and attains equality in Eq.~\eqref{eq:conditional_kl_penalty_identity}, so the primal maximum equals the dual value at $\eta^*$. The same identity shows that this primal maximizer is unique. Its KL constraint is active.

For $\rho=0$, the only feasible distribution is $u$, and
\begin{equation}
\widehat R_g^{\mathrm{rob}}(\theta)
=\frac{1}{n_g}\sum_{j\in I_g}L_j(\theta).
\label{eq:conditional_kl_rho_zero}
\end{equation}
Taking $\eta\to\infty$ in the dual objective recovers this mean. For $\rho\geq\rho_{\mathrm{sat}}$, the uniform distribution on $M$ is feasible and attains $\max_jL_j$. The dual objective has the same limiting value as $\eta\downarrow0$, because
\begin{equation}
\lim_{\eta\downarrow0}\eta\log\!\left(
\frac{1}{n}\sum_j e^{L_j/\eta}\right)=\max_jL_j.
\label{eq:conditional_kl_saturation_limit}
\end{equation}
If all losses equal a constant $c$, every feasible distribution has expected loss $c$, and the dual expression is $c+\eta\rho$; its infimum is also $c$. This includes $n_g=1$. These cases complete the derivation of Eq.~\eqref{eq:conditional_kl_dual} for every nonnegative radius.

For an interior radius and its finite optimizer, higher-loss patients have larger weights: $q_j^{(g)}(\eta_g^*)/q_k^{(g)}(\eta_g^*)=\exp((L_j-L_k)/\eta_g^*)$. More generally, any maximizing vector satisfies $L_j>L_k\Rightarrow q_j^*\geq q_k^*$. Otherwise, swapping those two weights would preserve their KL divergence under the uniform nominal distribution and strictly increase expected loss. At a saturated radius, distinct nonmaximal losses can both receive zero weight, so the general inequality is non-strict. Uniform weights are feasible for every radius, giving
\begin{equation}
\widehat R_g^{\mathrm{mean}}(\theta)
\leq\widehat R_g^{\mathrm{rob}}(\theta)
\leq\max_{j\in I_g}L_j(\theta),
\label{eq:conditional_kl_risk_bounds}
\end{equation}
and the robust risk is nondecreasing in $\rho_g$ because its feasible sets are nested.

\paragraph{Mean-loss baseline and subgroup weighting.}
At zero radii, Eq.~\eqref{eq:duetmoeplus_group_risk} becomes the sum of subgroup means,
\begin{equation}
\mathcal L_{\mathrm{mean}}(\theta)
=\sum_{g\in\mathcal G}\widehat R_g^{\mathrm{mean}}(\theta)
=\sum_{g\in\mathcal G}\frac{1}{n_g}
\sum_{j\in I_g}L_j(\theta).
\label{eq:conditional_kl_group_balanced_mean}
\end{equation}
Dividing this objective by $|\mathcal G|$ gives the group-balanced empirical risk. In contrast, with $N=\sum_g n_g$, the pooled empirical risk is
\begin{equation}
\widehat R_{\mathrm{pool}}(\theta)
=\frac{1}{N}\sum_jL_j(\theta)
=\sum_{g\in\mathcal G}\frac{n_g}{N}
\widehat R_g^{\mathrm{mean}}(\theta).
\label{eq:conditional_kl_pooled_mean}
\end{equation}
The group-balanced and pooled objectives assign different relative subgroup weights when subgroup sizes differ. Accordingly, the mean-loss version in Proposition~\ref{prop:duetmoeplus_variance} preserves the subgroup aggregation of Eq.~\eqref{eq:duetmoeplus_group_risk}; identifying an empirical baseline with this zero-radius case requires the same aggregation and patient-loss definition.

\paragraph{Gradient interpretation.}
Assume the patient losses are continuously differentiable near $\theta$. For a realized stochastic forward pass, hold the underlying random draws fixed. With top-$K$ routing and ReLU activations, the following statement applies on regions where the stated differentiability assumption holds. Write
$\mathcal C_g=\{q\in\Delta_{n_g}:D_{\mathrm{KL}}(q\Vert u_g)\leq\rho_g\}$ and
$\mathcal S_g(\theta)=\operatorname*{arg\,max}_{q\in\mathcal C_g}\sum_jq_jL_j(\theta)$.
For a direction $v$, the one-sided directional derivative is
\begin{equation}
\bigl(\widehat R_g^{\mathrm{rob}}\bigr)'(\theta;v)
=\max_{q\in\mathcal S_g(\theta)}
\sum_{j\in I_g}q_j\nabla_\theta L_j(\theta)^\top v.
\label{eq:conditional_kl_directional_derivative}
\end{equation}
To see this, set $F_g(\theta,q)=\sum_jq_jL_j(\theta)$ and choose an optimizer $q_h$ at $\theta+hv$, $h>0$. For any $q\in\mathcal S_g(\theta)$, optimality gives
\begin{equation}
\begin{aligned}
F_g(\theta+hv,q)-F_g(\theta,q)
&\leq\widehat R_g^{\mathrm{rob}}(\theta+hv)
-\widehat R_g^{\mathrm{rob}}(\theta)\\
&\leq F_g(\theta+hv,q_h)-F_g(\theta,q_h).
\end{aligned}
\label{eq:conditional_kl_envelope_bounds}
\end{equation}
Compactness and continuity imply that every subsequential limit of $q_h$ belongs to $\mathcal S_g(\theta)$. The first-order Taylor remainder is uniform over $\mathcal C_g$, since finitely many patient losses are combined with simplex weights. Dividing these bounds by $h$ gives Eq.~\eqref{eq:conditional_kl_directional_derivative}. If $\mathcal S_g(\theta)=\{q^{(g),*}\}$, the same argument for arbitrary parameter perturbations establishes differentiability and
\begin{equation}
\nabla_\theta\widehat R_g^{\mathrm{rob}}(\theta)
=\sum_{j\in I_g}q_j^{(g),*}\nabla_\theta L_j(\theta).
\label{eq:conditional_kl_gradient}
\end{equation}
When every subgroup maximizer is unique, the gradient of the sum in Eq.~\eqref{eq:duetmoeplus_group_risk} is the sum of these weighted subgroup gradients. This envelope calculation accounts for the loss dependence of the optimal weights; it does not require convexity of the neural-network losses in $\theta$. At a nonunique maximizer, Eq.~\eqref{eq:conditional_kl_directional_derivative} applies without asserting an ordinary gradient.

\subsubsection{Proof of Proposition~\ref{prop:duetmoeplus_variance}}
\label{app:conditional_kl_proof}

\begin{proof}
Fix $\theta$ and a nonempty subgroup $g$. Define the moments under its uniform empirical distribution by
\begin{equation}
\mu_g=\frac{1}{n_g}\sum_{j\in I_g}L_j(\theta),
\qquad z_j=L_j(\theta)-\mu_g,
\qquad \sigma_g^2=\frac{1}{n_g}\sum_{j\in I_g}z_j^2.
\label{eq:conditional_kl_empirical_moments}
\end{equation}
Thus, the standard deviation uses the empirical-distribution denominator $n_g$, rather than the unbiased-estimation denominator $n_g-1$. The recovery of the mean-loss objective when all radii are zero follows from Eq.~\eqref{eq:conditional_kl_rho_zero}.

If $\sigma_g=0$, all $z_j$ equal zero. Every feasible distribution then has expected loss $\mu_g$, so $\widehat R_g^{\mathrm{rob}}(\theta)=\mu_g$ for every $\rho_g\geq0$. This also proves the claimed expansion in this case, with zero remainder.

Suppose $\sigma_g>0$, and temporarily omit $g$ from $n$, $\mu$, $\sigma$, and $\rho$. Define the centered log moment-generating function
\begin{equation}
\psi(t)=\log\!\left(\frac{1}{n}\sum_j e^{tz_j}\right).
\label{eq:conditional_kl_centered_cgf}
\end{equation}
The finite loss vector makes $\psi$ smooth near zero, with
$\psi(0)=\psi'(0)=0$ and $\psi''(0)=\sigma^2$. Taylor expansion therefore gives
\begin{equation}
\psi(t)=\frac{\sigma^2}{2}t^2+O(t^3),
\qquad \psi'(t)=\sigma^2t+O(t^2).
\label{eq:conditional_kl_cgf_expansion}
\end{equation}
For all sufficiently small positive $\rho$, the preceding KL derivation provides a unique positive tilt $t_\rho$ satisfying
\begin{equation}
\rho=t_\rho\psi'(t_\rho)-\psi(t_\rho)
=\frac{\sigma^2}{2}t_\rho^2+O(t_\rho^3),
\qquad
\widehat R_g^{\mathrm{rob}}(\theta)=\mu+\psi'(t_\rho).
\label{eq:conditional_kl_small_radius_tilt}
\end{equation}
Centering does not change the tilt weights or their KL divergence. Moreover, $t_\rho\to0$ as $\rho\downarrow0$, since the divergence is continuous and strictly increasing from zero.

The positive quadratic coefficient in Eq.~\eqref{eq:conditional_kl_small_radius_tilt} implies $t_\rho=O(\sqrt\rho)$. Hence
$t_\rho^2-2\rho/\sigma^2=O(\rho^{3/2})$. Dividing this difference of squares by
$t_\rho+\sqrt{2\rho}/\sigma\geq\sqrt{2\rho}/\sigma$ yields
\begin{equation}
t_\rho=\frac{\sqrt{2\rho}}{\sigma}+O(\rho).
\label{eq:conditional_kl_tilt_expansion}
\end{equation}
Substituting into the risk identity in Eq.~\eqref{eq:conditional_kl_small_radius_tilt} gives
\begin{equation}
\widehat R_g^{\mathrm{rob}}(\theta)
=\mu+\sigma^2t_\rho+O(t_\rho^2)
=\mu+\sqrt{2\rho}\,\sigma+O(\rho).
\label{eq:conditional_kl_variance_expansion}
\end{equation}
Restoring the subgroup index and using $\mu_g=\widehat R_g^{\mathrm{mean}}(\theta)$ proves Proposition~\ref{prop:duetmoeplus_variance}. The remainder is pointwise for fixed $g$ and $\theta$; its constant may depend on their finite loss vector.
\end{proof}

The expansion identifies a local standard-deviation term in the empirical robust risk. It describes the training objective, rather than guaranteeing reduced prediction variability or improved performance for every patient.

\section{Implementation Details}
\label{app:implementation_details}
\subsection{Dataset cohorts and splits}
\label{app:dataset_cohorts}
We report each protocol separately. The two RT evaluations use different cohorts and subgroup definitions, so their results are compared within each protocol.

\paragraph{PI-CAI (institution-grouped multimodal protocol).}
The cohort comprises the $1{,}500$ public PI-CAI cases (RUMC $800$, ZGT
$350$, PCNN $350$), each with T2-weighted, apparent diffusion coefficient,
and high-$b$-value MRI, a case-level csPCa label, and the clinical fields
listed in Section~\ref{sec:exp_setup}. We hold out one fold of a
patient-level five-fold split, stratified by institution and csPCa status,
as the test set ($301$ cases from $295$ patients: RUMC $160$, ZGT $70$,
PCNN $71$) and split the remaining $1{,}199$ cases at the patient level
into $960$ training and $239$ validation cases, so that no patient appears
in more than one split. Institution is the subgroup label for routing, and
the linked clinical fields form the clinical memory of the multimodal
models.

\paragraph{Harvard-FairSeg (race-grouped unimodal protocol).}
The official test split contains $2{,}000$ images: Asian $169$, Black $299$, and White $1{,}532$. The main and paired lower-tail comparisons use a shared KL radius $\rho_g=0.1$ across racial subgroups; Appendix~\ref{app:rho_selection} examines radius sensitivity.

\paragraph{RT (institution-grouped multimodal protocol).}
The fixed split contains $709$ training cases (SC $659$, YI $24$, GN $26$),
$105$ validation cases (SC $95$, YI $5$, GN $5$), and $420$ test cases
(SC $189$, YI $112$, GN $119$). Institution selects the router, and the
linked clinical fields listed in Section~\ref{sec:exp_setup} form the
clinical memory. These fields include a de-identified physician code, because target delineation can differ between physicians. The code is assigned per institution from the training split: physicians with at least five training cases receive integer codes in order of frequency, and all other physicians, including those absent from the training split or missing from the record, receive the reserved code $0$. At inference, an unknown or unrecorded physician is therefore represented by the same code as infrequent physicians during training, and no physician names are used.

\paragraph{RT (T-stage-grouped unimodal protocol).}
The image-only evaluation contains $275$ held-out cases: T1 $11$, T2 $129$, T3 $114$, and T4 $21$. The $275$ test cases are not treated as a subset of the $420$-case test split above.

\subsection{Fairness metrics}
\label{app:fairness_metrics}
\paragraph{Segmentation overlap.}
Tables~\ref{tab:rt_tstage_full} and~\ref{tab_optic} supplement the main Dice comparisons with intersection over union (IoU). For predicted and reference foreground masks $P$ and $Y$, with nonzero denominators,
\begin{equation}
\operatorname{Dice}(P,Y)=\frac{2|P\cap Y|}{|P|+|Y|},\qquad
\operatorname{IoU}(P,Y)=\frac{|P\cap Y|}{|P\cup Y|}.
\label{eq:segmentation_overlap_metrics}
\end{equation}
Both are higher-is-better scores. We compute them per case and average over the full evaluation set or each subgroup; FairSeg rim and cup are evaluated separately.

\paragraph{Equity-scaled performance.}
Let $M_{\mathrm{all}}$ and $M_g$ denote performance on the full evaluation set and within subgroup $g\in\mathcal G$. We report subgroup values, the worst-subgroup value $\min_g M_g$, and the equity-scaled score~\citep{tianfairseg}
\begin{equation}
\operatorname{ES}(M)=\frac{M_{\mathrm{all}}}{1+\sum_{g\in\mathcal G}|M_{\mathrm{all}}-M_g|}.
\label{eq:equity_scaled_metric}
\end{equation}
Mean Dice and IoU give ES-Dice and ES-IoU. The PI score averages case-level AUROC and lesion-level average precision~\citep{saha2024picai}; applying the same scaling gives ES-PI. These scores assess inter-subgroup performance and are complemented by the within-subgroup outcomes below.

\paragraph{Intra-subgroup outcomes.}
For PI-CAI, institution-level case-score ROC curves give sensitivity at $90\%$ specificity and partial AUROC over false-positive rates up to $0.2$ (Tables~\ref{tab:picai_sens_spec90} and~\ref{tab:picai_pauc}). For RT, we report patient-level Dice distributions and the 25th percentile within each T-stage subgroup. For FairSeg, the lowest-$10\%$ mean Dice averages the lowest $\lceil0.1n_g\rceil$ scores in subgroup $g$, ranked separately by method and target. Tail membership may therefore differ across methods. The radius analysis also reports the minimum of these subgroup lower-tail means.

\paragraph{Uncertainty.}
For methods with case-level predictions, $95\%$ CIs follow the $1{,}000$-resample subgroup bootstrap in Section~\ref{sec:exp_setup}; aggregate-only published results are shown as reported. CIs condition on the evaluated checkpoints and do not capture variation across training seeds or checkpoint selection.

\subsection{Experimental setup}
\label{app:experiment_setup}

\paragraph{Main 2D protocol: Harvard-FairSeg.}
The main 2D setup uses TransUNet~\citep{chen2021transunet} with a ViT-B encoder, following the published fairness-baseline protocol. Images are center-cropped and resized to $224\times224$ pixels. Training uses a batch size of $42$, an initial learning rate of $1\times10^{-2}$, and $100$ epochs; the reported checkpoint is selected among those saved every $10$ epochs. The lower-tail comparison in Figure~\ref{fig:fairseg_tail} and radius sweep in Appendix~\ref{app:rho_selection} use a separate $80$-epoch schedule with the fixed final checkpoint (epoch $79$).

\paragraph{Main 3D protocol: image-only RT.}
The image-only RT setup describes a 3D residual U-Net backbone based on 3D U-Net~\citep{cciccek20163d}, as used for radiotherapy target delineation~\citep{oh2023llm}. Training samples $384\times384\times128$-voxel patches with a batch size of $4$, and evaluation applies sliding-window inference to full CT volumes. The training schedule runs for up to $100$ epochs, starting at $5\times10^{-5}$, with validation-driven early stopping.

\paragraph{Software and hardware.}
The 2D and 3D implementations report PyTorch~\citep{paszke2019pytorch} on CUDA 11.8, AdamW~\citep{loshchilov2017decoupled} with exponential learning-rate decay, a single NVIDIA A100 80\,GB GPU for 2D, and a single NVIDIA RTX A6000 48\,GB GPU for 3D. 

\paragraph{Multimodal protocols.}
The segmentation networks {in the main multimodal comparisons} are trained from scratch, while text encoders are frozen pretrained models; training uses AdamW{~\citep{loshchilov2017decoupled}} for $101$ epochs,
with $10$ warm-up epochs followed by cosine decay and weight decay
$10^{-5}$; the validation split is used only for monitoring, and {these comparisons}
use the checkpoint after the final epoch. The base learning
rate is $5\times10^{-5}$ for radiotherapy (batches of four 3D crops) and
$8.66\times10^{-5}$ for PI-CAI (batches of $12$ cases, scaled by the square
root of the batch-size ratio), and {DuetMoE's six pathway coefficients at each clinical-context stage} use
$16\times$ the base learning rate. Radiotherapy uses cross-entropy plus
foreground soft Dice loss. PI-CAI uses cross-entropy plus softmax focal loss
with equal weights and deep supervision with weights $(8,4,2,1)/15$; {for PI-CAI, each}
epoch consists of $80$ optimizer steps, and each batch draws two
cancer-positive and two cancer-negative cases from each of the three
institutions, {giving equal case counts per institution at each
update.} {The main multimodal runs} use a single NVIDIA RTX PRO 6000 Blackwell Server
Edition GPU.

\section{Additional Experimental Results}

\subsection{\texorpdfstring{{PI-CAI: intra-subgroup detection performance}}{PI-CAI: intra-subgroup detection performance}} \label{picai-intra}

{Tables~\ref{tab:picai_sens_spec90} and~\ref{tab:picai_pauc} complement the PI scores in Table~\ref{tab:picai_multimodal} with detection performance at high specificity within institutions. \textbf{DuetMoE} raises RUMC sensitivity from $0.574$ with the strongest baseline for this metric to $0.681$ ($+10.7$ percentage points) and achieves the highest normalized partial AUROC in RUMC and ZGT.}

\begin{table}[h!]
\centering
\caption{{\textbf{PI-CAI.} Sensitivity at $90\%$ specificity for \textbf{multimodal} csPCa detection, evaluated overall and separately within each institution from case scores. Here $n_{+}$ is the number of cancer-positive cases.}}
\label{tab:picai_sens_spec90}
\resizebox{\linewidth}{!}{%
\begin{tabular}{lcccc}
\toprule
Method & All ($n_{+}=85$) & RUMC ($n_{+}=47$) & ZGT ($n_{+}=16$) & PCNN ($n_{+}=22$) \\
\midrule
3D ResUNet & 0.459 & 0.468 & {\underline{0.625}} & \textbf{0.364} \\
ConTEXTualNet3D-based & 0.435 & 0.532 & 0.438 & 0.136 \\
PCSA-Seg & 0.424 & 0.532 & 0.375 & 0.273 \\
dMoE & {\underline{0.518}} & {\underline{0.574}} & \textbf{0.688} & {\underline{0.318}} \\
MoME & 0.506 & 0.553 & {\underline{0.625}} & {\underline{0.318}} \\
\textbf{DuetMoE} & \textbf{0.565} & \textbf{0.681} & \textbf{0.688} & 0.182 \\
\bottomrule
\end{tabular}%
}
\par\smallskip
{\raggedright\footnotesize{Bold and underlined point estimates indicate the highest and second-highest values in each column, respectively; ties share the same marking.}\par}
\end{table}

\begin{table}[h!]
\centering
\caption{{\textbf{PI-CAI.} Normalized partial AUROC over false-positive rates up to $0.2$ for \textbf{multimodal} csPCa detection, evaluated overall and separately within each institution from case scores.}}
\label{tab:picai_pauc}
\resizebox{\linewidth}{!}{%
\begin{tabular}{lcccc}
\toprule
Method & All ($n=301$) & RUMC ($n=160$) & ZGT ($n=70$) & PCNN ($n=71$) \\
\midrule
3D ResUNet & 0.447 & 0.409 & 0.605 & \textbf{0.353} \\
{ConTEXTualNet3D-based} & 0.394 & 0.473 & 0.416 & 0.202 \\
PCSA-Seg & 0.407 & 0.494 & 0.433 & 0.252 \\
dMoE & 0.459 & 0.500 & {\underline{0.635}} & 0.238 \\
MoME & \textbf{0.500} & {\underline{0.546}} & 0.605 & {\underline{0.311}} \\
\textbf{DuetMoE} & {\underline{0.488}} & \textbf{0.564} & \textbf{0.670} & 0.185 \\
\bottomrule
\end{tabular}%
}
\par\smallskip
{\raggedright\footnotesize{Bold and underlined point estimates indicate the highest and second-highest values in each column, respectively; ties share the same marking.}\par}
\end{table}

\subsection{\texorpdfstring{{Unimodal radiotherapy: additional evaluation metrics}}{Unimodal radiotherapy: additional evaluation metrics}} \label{app:rt_tstage_supplement}

{Table~\ref{tab:rt_tstage_full} extends the Dice comparison in Table~\ref{tab:rt_tstage} with IoU and ES-IoU. \textbf{DuetMoE+} also achieves the highest ES-IoU ($0.411$) and raises IoU in T2, the lowest-scoring stage for dMoE, from $0.435$ to $0.473$, supporting improved inter-subgroup performance.}

\begin{table*}[h!]
\centering
\caption{{\textbf{Radiotherapy.} Extended comparison on \textbf{unimodal} 3D target segmentation using tumor stage as the subgroup attribute, supplementing Table~\ref{tab:rt_tstage} with IoU and ES-IoU.}}
\label{tab:rt_tstage_full}
\resizebox{\linewidth}{!}{%
\setlength{\tabcolsep}{3pt}%
\begin{tabular}{lcccccccccccc}
\toprule
Method & \multicolumn{4}{c}{All ($n=275$)} & \multicolumn{2}{c}{T1 ($n=11$)} & \multicolumn{2}{c}{T2 ($n=129$)} & \multicolumn{2}{c}{T3 ($n=114$)} & \multicolumn{2}{c}{T4 ($n=21$)} \\
\cmidrule(l){2-5} \cmidrule(lr){6-7} \cmidrule(lr){8-9} \cmidrule(lr){10-11} \cmidrule(lr){12-13}
 & {ES-Dice ($95\%$ CI)} & {Dice ($95\%$ CI)} & {ES-IoU ($95\%$ CI)} & {IoU ($95\%$ CI)} & Dice & IoU & Dice & IoU & Dice & IoU & Dice & IoU \\
\midrule
3D ResUNet \cite{cciccek20163d} & 0.487 (0.447--0.529) & 0.610 (0.589--0.630) & 0.367 (0.336--0.399) & 0.462 (0.440--0.482) & {0.492} & {0.341} & {0.570} & 0.420 & 0.659 & 0.511 & 0.656 & 0.506 \\
+ FEBS \cite{tianfairseg} & 0.434 (0.406--0.467) & 0.586 (0.567--0.604) & 0.322 (0.302--0.346) & 0.433 (0.414--0.452) & {0.466} & {0.288} & {0.522} & 0.374 & 0.652 & 0.501 & {0.688} & 0.527 \\
+ MoE \cite{shazeer2017outrageously} & 0.452 (0.415--0.492) & 0.608 (0.586--0.628) & 0.342 (0.314--0.372) & 0.461 (0.439--0.482) & {0.492} & {0.345} & {0.543} & 0.393 & 0.674 & 0.532 & 0.708 & 0.557 \\
+ GDRO \cite{sagawa2019distributionally} & {0.454 (0.421--0.491)} & {0.607 (0.586--0.627)} & 0.340 (0.314--0.372) & 0.459 (0.436--0.481) & {0.481} & {0.329} & {0.546} & 0.395 & {0.671} & 0.528 & {0.694} & 0.547 \\
+ Prompt-GDRO \cite{panaganti2026groupd} & {0.476 (0.440--0.517)} & {0.612 (0.592--0.631)} & 0.357 (0.329--0.392) & 0.462 (0.441--0.482) & {0.492} & {0.338} & 0.562 & 0.411 & 0.669 & 0.523 & {0.673} & 0.520 \\
+ dMoE \cite{oh2025distribution} & {\underline{0.499} (0.469--0.531)} & {\underline{0.650} (0.628--0.671)} & {\underline{0.384} (0.351--0.419)} & {\underline{0.507} (0.484--0.528)} & {\underline{0.718}} & {\underline{0.571}} & {\underline{0.586}} & \underline{0.435} & \textbf{0.693} & \textbf{0.556} & \textbf{0.778} & \textbf{0.641} \\
+ \textbf{DuetMoE+} & {\textbf{0.530} (0.499--0.569)} & {\textbf{0.665} (0.643--0.683)} & {\textbf{0.411} (0.384--0.444)} & {\textbf{0.521} (0.498--0.540)} & {\textbf{0.757}} & \textbf{0.621} & {\textbf{0.621}} & {\textbf{0.473}} & {\underline{0.688}} & {\underline{0.547}} & {\underline{0.758}} & {\underline{0.615}} \\
\bottomrule
\end{tabular}}
\par\smallskip

{\raggedright\footnotesize
{Bold and underlined point estimates indicate the highest and second-highest values in each column, respectively; ties share the same marking. Parentheses show $95\%$ confidence intervals.}\par}
\end{table*}

\subsection{\texorpdfstring{{Harvard-FairSeg: subgroup performance and patient lower tails}}{Harvard-FairSeg: subgroup performance and patient lower tails}} \label{app:fairseg-rim}

\begin{figure*}[t]
\centering
\includegraphics[width=0.82\textwidth]{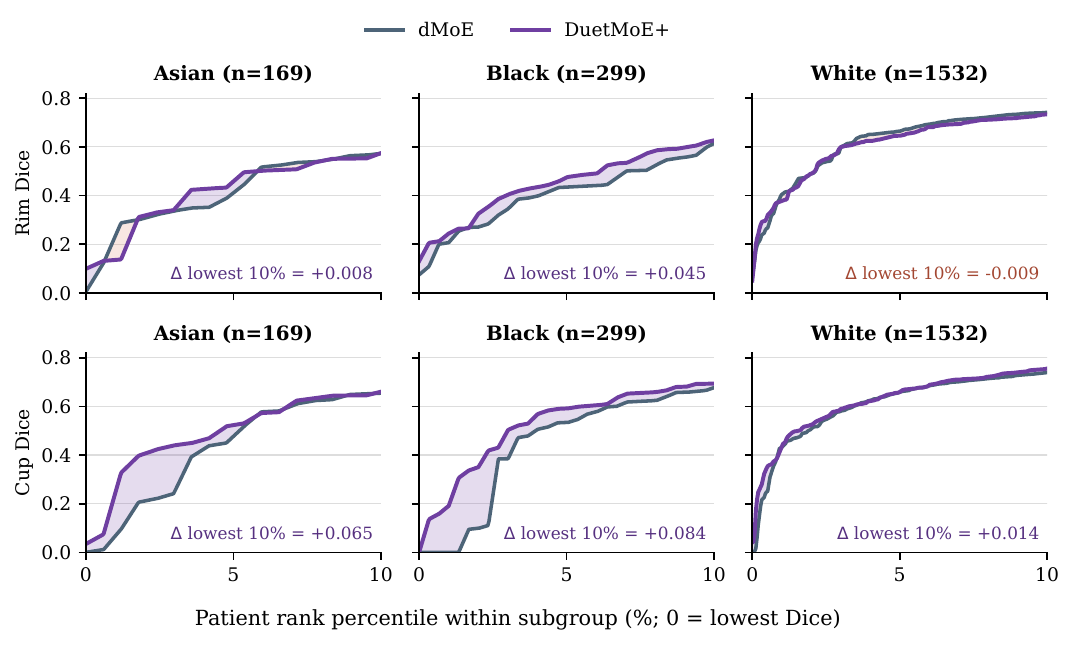}
\caption{{\textbf{Harvard-FairSeg: intra-subgroup performance.} Dice curves for the lowest-scoring $10\%$ of patients within each racial subgroup, comparing dMoE and \textbf{DuetMoE+}. \textbf{Top:} neuroretinal rim; \textbf{bottom:} optic cup. The horizontal axis is patient rank percentile ($0$ denotes the lowest Dice); patients are ranked separately by method, subgroup, and target. Each $\Delta$ is the difference in lowest-$10\%$ mean Dice (\textbf{DuetMoE+} minus dMoE).}}
\label{fig:fairseg_tail}
\end{figure*}

{Table~\ref{tab_optic} extends the cup Dice comparison in Table~\ref{tab:fairseg-cup} to both segmentation targets and additional overlap metrics. \textbf{DuetMoE+} achieves the highest ES-Dice for both neuroretinal rim ($0.755$) and optic cup ($0.843$), providing further evidence on inter-subgroup performance. Figure~\ref{fig:fairseg_tail} complements these subgroup scores with the lowest-$10\%$ patient Dice within each racial subgroup.}

\begin{table*}[h!]
\centering
\caption{{\textbf{Harvard-FairSeg.} Extended comparison on \textbf{unimodal} 2D neuroretinal-rim and optic-cup segmentation using race as the subgroup attribute. Dice, IoU, and their equity-scaled scores supplement the cup Dice results in Table~\ref{tab:fairseg-cup}.}}
\label{tab_optic}
\resizebox{\linewidth}{!}{%
\begin{tabular}{lcccccccccc}
\toprule
Method & \multicolumn{4}{c}{All ($n=2000$)} & \multicolumn{2}{c}{Asian ($n=169$)} & \multicolumn{2}{c}{Black ($n=299$)} & \multicolumn{2}{c}{White ($n=1532$)} \\
\cmidrule(l){2-5} \cmidrule(lr){6-7} \cmidrule(lr){8-9} \cmidrule(lr){10-11}
 & {ES-Dice ($95\%$ CI)} & {Dice ($95\%$ CI)} & {ES-IoU ($95\%$ CI)} & {IoU ($95\%$ CI)} & Dice & IoU & Dice & IoU & Dice & IoU \\
\midrule
\multicolumn{11}{l}{{\textbf{Neuroretinal rim}}} \\
\cmidrule(l){1-11}
TransUNet$^\dagger$ \cite{chen2021transunet} & 0.703 & 0.793 & 0.585 & 0.671 & 0.746 & 0.616 & {0.731} & {0.599} & 0.811 & 0.691 \\
+ ADV$^\dagger$ \cite{madras2018learning} & 0.700 & 0.791 & 0.583 & 0.668 & 0.741 & 0.612 & {0.729} & {0.598} & 0.809 & 0.689 \\
+ FEBS$^\dagger$ \cite{tianfairseg} & 0.705 & 0.795 & 0.587 & 0.673 & 0.748 & 0.619 & {0.733} & {0.602} & 0.813 & 0.694 \\
+ FairDiff$^\ddagger$ \cite{li2024fairdiff} & 0.716 & 0.800 & 0.596 & 0.680 & 0.757 & 0.628 & {0.743} & {0.611} & 0.817 & 0.699 \\
+ MoE \cite{shazeer2017outrageously} & 0.733 (0.713--0.752) & 0.804 (0.799--0.809) & 0.614 (0.596--0.633) & 0.685 (0.680--0.691) & {0.760} & {0.635} & 0.763 & 0.635 & 0.817 & 0.701 \\
+ GDRO \cite{sagawa2019distributionally} & 0.742 (0.722--0.763) & 0.809 (0.804--0.814) & 0.625 (0.607--0.645) & 0.691 (0.686--0.697) & {0.764} & {0.641} & 0.775 & 0.649 & 0.820 & 0.705 \\
+ Prompt-GDRO \cite{panaganti2026groupd} & {\underline{0.747} (0.726--0.767)} & {\textbf{0.817} (0.812--0.822)} & {\underline{0.631} (0.612--0.649)} & {\textbf{0.703} (0.697--0.708)} & {0.767} & {0.643} & \textbf{0.784} & \textbf{0.663} & \textbf{0.828} & \textbf{0.717} \\
+ dMoE \cite{oh2025distribution} & 0.743 (0.723--0.763) & 0.813 (0.808--0.818) & 0.627 (0.608--0.645) & 0.698 (0.692--0.704) & \underline{0.769} & \underline{0.645} & 0.776 & 0.652 & 0.825 & 0.713 \\
+ \textbf{DuetMoE+} & {\textbf{0.755} (0.735--0.775)} & {\underline{0.816} (0.811--0.821)} & {\textbf{0.639} (0.619--0.657)} & {\underline{0.701} (0.696--0.707)} & {\textbf{0.780}} & {\textbf{0.658}} & {\underline{0.782}} & {\underline{0.659}} & {\underline{0.827}} & {\underline{0.714}} \\
\cmidrule(l){1-11}
\multicolumn{11}{l}{{\textbf{Optic cup}}} \\
\cmidrule(l){1-11}
TransUNet$^\dagger$ \cite{chen2021transunet} & 0.828 & 0.848 & 0.730 & 0.753 & {0.827} & {0.728} & 0.849 & 0.758 & 0.850 & 0.755 \\
+ ADV$^\dagger$ \cite{madras2018learning} & 0.826 & 0.841 & 0.727 & 0.743 & {0.825} & {0.726} & 0.842 & 0.748 & 0.843 & 0.744 \\
+ FEBS$^\dagger$ \cite{tianfairseg} & 0.825 & 0.846 & 0.727 & 0.750 & {0.825} & {0.725} & 0.848 & 0.755 & 0.848 & 0.751 \\
+ FairDiff$^\ddagger$ \cite{li2024fairdiff} & {0.833} & 0.848 & 0.736 & 0.753 & {0.832} & {0.735} & 0.848 & 0.757 & 0.850 & 0.754 \\
+ MoE \cite{shazeer2017outrageously} & 0.830 (0.809--0.847) & 0.854 (0.849--0.860) & 0.739 (0.720--0.754) & 0.762 (0.755--0.768) & 0.845 & 0.757 & {0.842} & {0.748} & 0.857 & 0.765 \\
+ GDRO \cite{sagawa2019distributionally} & 0.840 (0.818--0.857) & 0.860 (0.855--0.865) & {\underline{0.750} (0.728--0.766)} & 0.769 (0.763--0.775) & \underline{0.850} & 0.760 & 0.851 & {0.759} & 0.863 & 0.772 \\
+ Prompt-GDRO \cite{panaganti2026groupd} & {\underline{0.842} (0.819--0.861)} & {\textbf{0.866} (0.861--0.871)} & {\textbf{0.755} (0.732--0.775)} & {\textbf{0.778} (0.772--0.785)} & {0.849} & \underline{0.761} & \textbf{0.859} & \textbf{0.770} & \textbf{0.870} & \textbf{0.782} \\
+ dMoE \cite{oh2025distribution} & 0.832 (0.810--0.853) & 0.862 (0.856--0.867) & 0.745 (0.722--0.765) & 0.773 (0.766--0.779) & {0.844} & {0.755} & 0.851 & 0.761 & 0.866 & 0.777 \\
+ \textbf{DuetMoE+} & {\textbf{0.843} (0.822--0.861)} & {\underline{0.865} (0.860--0.869)} & {\textbf{0.755} (0.733--0.773)} & {\underline{0.775} (0.769--0.781)} & {\textbf{0.852}} & {\textbf{0.763}} & {\underline{0.856}} & {\underline{0.767}} & {\underline{0.868}} & {\underline{0.778}} \\
\bottomrule
\end{tabular}}
\par\smallskip

{\raggedright\footnotesize
{Bold and underlined point estimates indicate the highest and second-highest values within each target and column, respectively; ties share the same marking. Parentheses show $95\%$ confidence intervals.} $^\dagger$ Metrics reported in~\cite{tianfairseg}. $^\ddagger$ ES metrics recalculated from~\cite{li2024fairdiff} using Eq.~\eqref{eq:equity_scaled_metric}.\par}

\end{table*}

\section{Ablation}
\label{app:ablation}

\subsection{Clinical-context configurations}
\label{app:duetmoe_context_configurations}

We compare {clinical-context configurations} built from the reads defined in Section~\ref{sec:duetmoe}. Global-context configurations inject $G_j$ before routing or at the expert inputs; expert-specific clinical guidance injects $P_{j,e}$ at the expert inputs, optionally with the routing hint $B(A_j^G)$ added to the routing scores. {\textbf{DuetMoE} embeds global and spatial clinical reads before routing and expert-specific guidance afterward, with a memory-derived routing hint and learned pathway coefficients.} {Image-only MoE and multimodal dMoE provide reference configurations.} {These configurations examine where clinical memory enters routing and expert computation.}

{The radiotherapy configurations in Table~\ref{tab:duetmoe_ptv_context_legacy} use the main cohort split but an earlier protocol with a reduced clinical prompt and different memory construction and evaluation seeding. Their rows are compared within that protocol, so the DuetMoE score there ($0.742$) is not comparable with the main-table value ($0.761$ in Table~\ref{tab:rt_multimodal}). Table~\ref{tab:duetmoe_picai_context} evaluates PI-CAI configurations on the main test split.}

{\textbf{DuetMoE} achieves the highest overall Dice and ES-Dice in Table~\ref{tab:duetmoe_ptv_context_legacy}. On PI-CAI (Table~\ref{tab:duetmoe_picai_context}), it shares the highest overall PI score ($0.623$) with the expert-specific guidance configuration and achieves the highest ES-PI ($0.500$). These comparisons favor the full clinical-memory design for subgroup-aware learning.}

\begin{table}[h!]
\centering
\caption{{\textbf{Radiotherapy.} Clinical-context configurations for \textbf{multimodal} 3D target segmentation using institution as the subgroup attribute, evaluated under the earlier input and evaluation protocol described in the text.}}
\label{tab:duetmoe_ptv_context_legacy}
{\fontsize{8}{9.5}\selectfont
\setlength{\tabcolsep}{3pt}
\renewcommand{\arraystretch}{1.05}
\begin{tabular*}{\linewidth}{@{}l@{\extracolsep{\fill}}ccccc@{}}
\toprule
Configuration & \multicolumn{2}{c}{All} & SC & YI & GN \\
 & \multicolumn{2}{c}{($n=420$)} & ($n=189$) & ($n=112$) & ($n=119$) \\
\cmidrule(lr){2-3}\cmidrule(lr){4-4}\cmidrule(lr){5-5}\cmidrule(lr){6-6}
 & Dice & ES-Dice & Dice & Dice & Dice \\
\midrule
Image-only MoE & 0.680 & 0.593 & 0.727 & 0.694 & 0.594 \\
dMoE & 0.736 & 0.651 & 0.755 & 0.779 & 0.667 \\
Global clinical context at expert input & \underline{0.739} & \underline{0.660} & 0.757 & 0.777 & \underline{0.675} \\
Expert-specific clinical guidance & 0.734 & 0.644 & 0.757 & 0.775 & 0.659 \\
Expert-specific clinical guidance + routing hint & \underline{0.739} & 0.646 & \textbf{0.761} & \textbf{0.784} & 0.662 \\
\midrule
\textbf{DuetMoE} & \textbf{0.742} & \textbf{0.662} & \underline{0.760} & \underline{0.782} & \textbf{0.678} \\
\bottomrule
\end{tabular*}
}
\par\smallskip
{\raggedright\footnotesize {Bold and underlining mark the highest and second-highest distinct values per column; ties share the marking.}\par}
\end{table}

\begin{table}[h!]
\centering
\caption{{\textbf{PI-CAI.} Clinical-context configurations for \textbf{multimodal} csPCa detection using institution as the subgroup attribute, evaluated on the main test split.}}
\label{tab:duetmoe_picai_context}
{\fontsize{8}{9.5}\selectfont
\setlength{\tabcolsep}{3pt}
\renewcommand{\arraystretch}{1.05}
\begin{tabular*}{\linewidth}{@{}l@{\extracolsep{\fill}}ccccc@{}}
\toprule
Configuration & \multicolumn{2}{c}{All} & RUMC & ZGT & PCNN \\
 & \multicolumn{2}{c}{($n=301$)} & ($n=160$) & ($n=70$) & ($n=71$) \\
\cmidrule(lr){2-3}\cmidrule(lr){4-4}\cmidrule(lr){5-5}\cmidrule(lr){6-6}
 & PI score & ES-PI & PI score & PI score & PI score \\
\midrule
Image-only MoE & 0.595 & 0.469 & 0.651 & 0.648 & 0.436 \\
dMoE & 0.590 & 0.453 & 0.647 & \textbf{0.671} & 0.423 \\
Global clinical context before routing & 0.577 & 0.445 & 0.659 & 0.607 & 0.390 \\
Global clinical context at expert input & 0.592 & 0.479 & 0.649 & 0.620 & \underline{0.442} \\
Expert-specific clinical guidance & \textbf{0.623} & \underline{0.487} & \underline{0.686} & 0.634 & 0.418 \\
Expert-specific clinical guidance + routing hint & \underline{0.610} & 0.465 & 0.677 & \underline{0.660} & 0.414 \\
\midrule
\textbf{DuetMoE} & \textbf{0.623} & \textbf{0.500} & \textbf{0.688} & 0.637 & \textbf{0.457} \\
\bottomrule
\end{tabular*}
}
\par\smallskip
{\raggedright\footnotesize {Bold and underlining mark the highest and second-highest distinct values per column; ties share the marking.}\par}
\end{table}
\FloatBarrier

\subsection{Inter- and intra-subgroup components of DuetMoE+}
\label{app:ablation_two_axes}

We ablate inter-subgroup routing and intra-subgroup robust training in the unimodal \textbf{DuetMoE+}. \textbf{dMoE-ERM} retains subgroup-selected expert routing with ERM training. \textbf{Subgroup-DRO (SG-DRO)} applies the within-subgroup robust objective to a shared backbone without subgroup-selected routing. \textbf{DuetMoE+} embeds this objective in subgroup-routed expert learning.

\begin{table}[h!]
\centering
\caption{\textbf{Harvard-FairSeg and radiotherapy.} Ablation of subgroup-selected routing and within-subgroup robust training in \textbf{unimodal} segmentation, using race and tumor stage as the respective subgroup attributes.}
\label{tab:ablation_two_axes}
\resizebox{\linewidth}{!}{%
\begin{tabular}{llccccccc}
\toprule
Setting & Method & \multicolumn{4}{c}{Overall performance} & \multicolumn{2}{c}{Worst subgroup} & Component \\
\cmidrule(lr){3-6} \cmidrule(lr){7-8}
& & ES-Dice & Dice & ES-IoU & IoU & Dice & IoU & \\
\midrule
Harvard-FairSeg / Race (Rim)
& dMoE-ERM & 0.743 & 0.813 & 0.627 & 0.698 & 0.769 & 0.645 & Inter \\
& SG-DRO & \underline{0.749} & \underline{0.814} & \underline{0.633} & \underline{0.699} & \underline{0.774} & \underline{0.651} & Intra \\
& \textbf{DuetMoE+} & \textbf{0.755} & \textbf{0.816} & \textbf{0.639} & \textbf{0.701} & \textbf{0.780} & \textbf{0.658} & Inter + Intra \\
\midrule
Harvard-FairSeg / Race (Cup)
& dMoE-ERM & 0.832 & \underline{0.862} & 0.745 & \underline{0.773} & 0.844 & 0.755 & Inter \\
& SG-DRO & \underline{0.841} & 0.860 & \underline{0.752} & 0.769 & \underline{0.850} & \underline{0.759} & Intra \\
& \textbf{DuetMoE+} & \textbf{0.843} & \textbf{0.865} & \textbf{0.755} & \textbf{0.775} & \textbf{0.852} & \textbf{0.763} & Inter + Intra \\
\midrule
Radiotherapy / Tumor Stage
& dMoE-ERM & \underline{0.499} & \underline{0.650} & \underline{0.384} & \underline{0.507} & \underline{0.586} & \underline{0.435} & Inter \\
& SG-DRO & 0.490 & 0.616 & 0.370 & 0.468 & 0.515 & 0.359 & Intra \\
& \textbf{DuetMoE+} & \textbf{0.530} & \textbf{0.665} & \textbf{0.411} & \textbf{0.521} & \textbf{0.621} & \textbf{0.473} & Inter + Intra \\
\bottomrule
\end{tabular}%
}
\par\smallskip
{\raggedright\footnotesize Worst-subgroup scores are the minimum subgroup mean Dice or IoU across races (FairSeg) or tumor stages (RT). Component labels indicate the inter-subgroup routing and intra-subgroup robust-training mechanisms. Bold and underlined point estimates indicate the highest and second-highest distinct values within each setting and column, respectively; ties share the same marking.\par}
\end{table}
\FloatBarrier

Table~\ref{tab:ablation_two_axes} shows that the full \textbf{DuetMoE+} achieves the highest equity-scaled scores and worst-subgroup Dice and IoU in all three comparisons. These results support embedding within-subgroup robust training in subgroup-routed expert learning. The patient-level lower-tail analyses in Figures~\ref{fig:rt_tstage_analysis} and~\ref{fig:fairseg_tail} provide complementary evidence on intra-subgroup outcomes.

\subsection{KL-radius sensitivity and selection in DuetMoE+}
\label{app:rho_selection}

We examine sensitivity to the shared KL radius $\rho_g=\rho$ in \textbf{DuetMoE+} through a dedicated Harvard-FairSeg sweep. The main comparison uses $\rho=0.1$ (Appendix~\ref{app:dataset_cohorts}); the sweep varies the radius under a common training setup and fixed final-checkpoint rule. Table~\ref{tab:fairseg_rho_sensitivity} reports overall Dice, inter-subgroup metrics, and the minimum across subgroups of the lowest-$10\%$ mean Dice.

\begin{table*}[h!]
\centering
\caption{\textbf{Harvard-FairSeg.} KL-radius sensitivity of \textbf{DuetMoE+} on \textbf{unimodal} 2D neuroretinal-rim and optic-cup segmentation using race as the subgroup attribute. A shared radius $\rho_g=\rho$ is evaluated at the fixed final checkpoint.}
\label{tab:fairseg_rho_sensitivity}
\resizebox{\textwidth}{!}{%
\begin{tabular}{lcccccccc}
\toprule
& \multicolumn{4}{c}{Neuroretinal rim} & \multicolumn{4}{c}{Optic cup} \\
\cmidrule(lr){2-5}\cmidrule(lr){6-9}
$\rho$ & Dice & ES-Dice & Worst mean & Worst tail 10\% & Dice & ES-Dice & Worst mean & Worst tail 10\% \\
\midrule
0.01 & \textbf{0.818} & \textbf{0.753} & \textbf{0.777} & 0.392 & \textbf{0.863} & 0.840 & \textbf{0.851} & 0.481 \\
0.05 & \underline{0.817} & 0.750 & \textbf{0.777} & \underline{0.416} & 0.859 & 0.841 & 0.848 & 0.509 \\
0.10 & 0.815 & 0.743 & 0.766 & 0.363 & \underline{0.861} & 0.836 & 0.847 & 0.486 \\
0.15 & 0.815 & \underline{0.751} & \textbf{0.777} & \textbf{0.425} & \underline{0.861} & \underline{0.843} & \textbf{0.851} & \underline{0.512} \\
0.20 & 0.814 & 0.749 & 0.774 & 0.413 & 0.860 & 0.840 & \underline{0.849} & 0.496 \\
0.25 & 0.814 & 0.749 & \underline{0.775} & 0.410 & \underline{0.861} & \textbf{0.845} & \underline{0.849} & \textbf{0.524} \\
0.50 & 0.059 & 0.034 & 0.000 & 0.000 & 0.068 & 0.036 & 0.000 & 0.000 \\
\bottomrule
\end{tabular}}
\par\smallskip
{\raggedright\footnotesize Results use a single training seed, an $80$-epoch schedule, and the fixed final checkpoint on $2{,}000$ test images. Bold and underlining mark the highest and second-highest distinct values per target and column; ties share the marking. Worst mean is the minimum race-subgroup mean Dice. Worst tail 10\% is the minimum, across race subgroups, of the mean Dice among the $\lceil0.1|I_g|\rceil$ lowest-scoring patients in subgroup $g$. Patients are ranked separately for each radius, subgroup, and target.\par}
\end{table*}

Across the tested radii from $0.01$ to $0.25$, overall Dice varies little ($0.814$--$0.818$ for rim; $0.859$--$0.863$ for cup), while subgroup scores and lower tails vary more. Increasing $\rho$ from $0.10$ to $0.15$ raises the worst-subgroup lowest-$10\%$ mean Dice by $6.2$ and $2.6$ percentage points, respectively. This pattern motivates assessing radius choices with subgroup performance and patient lower tails alongside overall Dice.

\end{document}